\documentclass[letterpaper, 10 pt, conference]{ieeeconf}  %

\IEEEoverridecommandlockouts                              %

\usepackage[breaklinks=true,bookmarks=false]{hyperref}
\usepackage{graphicx,subcaption}
\usepackage{mathtools}
\usepackage{booktabs}
\usepackage{adjustbox}
\usepackage{multicol}
\usepackage{float}
\usepackage{multirow}
\usepackage{wrapfig}
\usepackage{bm}

\usepackage{amsmath,amssymb,stmaryrd,mathtools}
\usepackage{xcolor}
\usepackage{xspace}
\usepackage{bbm}
\usepackage{graphicx}

\definecolor{dark_red}{RGB}{122, 0, 0}
\definecolor{coral}{RGB}{255, 119, 94}
\definecolor{pink_orange}{RGB}{255, 72, 126}
\definecolor{vibrant_pink}{RGB}{255, 0, 104}
\definecolor{pink_pink}{RGB}{255, 37, 153}
\definecolor{wine}{RGB}{204, 0, 102}

\definecolor{light_orange}{RGB}{255, 198, 107}
\definecolor{orange(sae/ece)}{rgb}{1.0, 0.49, 0.0}
\definecolor{dark_orange}{RGB}{216,92,0}

\definecolor{org-purp-0}{RGB}{165, 76, 0}
\definecolor{org-purp-1}{RGB}{250, 130, 28}
\definecolor{org-purp-2}{RGB}{226, 89, 68}
\definecolor{org-purp-3}{RGB}{206, 92, 124}
\definecolor{org-purp-4}{RGB}{116, 80, 146}
\definecolor{org-purp-5}{RGB}{110, 78, 157}
\definecolor{org-purp-6}{RGB}{242, 177, 131}

\definecolor{teal(sae/ece)}{rgb}{0, 0.47, 0.52}
\definecolor{aqua}{RGB}{52,172,139}
\definecolor{dark_aqua}{RGB}{35,115,93}
\definecolor{dark_green}{RGB}{0, 92, 34}

\definecolor{grape}{RGB}{112,48,160}
\definecolor{purple}{rgb}{0.74, 0.65, 1.0}
\definecolor{dark_purple}{rgb}{0.58, 0.0, 0.82}
\definecolor{periwinkle}{RGB}{191, 140, 230}
\definecolor{purple-1}{RGB}{201, 121, 230}
\definecolor{purple-2}{RGB}{176,140,240} %

\definecolor{light_gray}{rgb}{0.9, 0.9, 0.9}
\definecolor{medium_gray}{rgb}{0.6, 0.6, 0.6} 
\definecolor{dark_gray}{rgb}{0.2, 0.2, 0.2} 
\definecolor{gray_1}{RGB}{100 ,100, 100}

\definecolor{sky_blue}{RGB}{37, 166, 213}
\definecolor{light_blue}{rgb}{0.33, 0.80, 1}
\definecolor{dark_blue}{rgb}{0.098, 0.239, 0.52}
\definecolor{ocean}{RGB}{13, 121, 202}
\definecolor{light_ocean}{RGB}{18, 178, 235}
\definecolor{dark_ocean}{RGB}{10, 89, 148}
\definecolor{vibrant_blue}{RGB}{14, 120, 255}

\definecolor{dark_brown}{rgb}{0.3255, 0.004, 0.001}

\newcommand{\para}[1]{\medskip\noindent\textbf{#1. }}

\newcounter{qnum}
\newcounter{tnum}
\newcommand{\acronym}{SEAL}
\newcommand{\ours}{\textcolor{org-purp-1}{\textbf{SEAL}}\xspace}
\DeclareRobustCommand{\reasonvla}{\textcolor{gray_1}{\textbf{\boldmath $\bm{\pi_0}$}-reason}\xspace}
\newcommand{\vgps}{\textcolor{purple-1}{\textbf{\boldmath{$\bm{\pi_0}$}-V-GPS}}\xspace}
\newcommand{\vla}{\textcolor{purple-2}{\textbf{\boldmath{$\bm{\pi_0}$}}}\xspace}

\newcommand{\obs}{o}

\newcommand{\lang}{\ell}

\newcommand{\langg}{\ell^g}
\newcommand{\langr}{\ell^r}
\newcommand{\langropt}{\ell^{r}}
\newcommand{\langrhat}{\hat{\ell}^r}

\newcommand{\oseq}{\mathbf{o}}

\newcommand{\traj}{\tau}
\newcommand{\act}{a}
\newcommand{\aseq}{\mathbf{\act}}

\newcommand{\aseqhat}{\mathbf{\hat{\act}}}

\usepackage{amsmath}

\title{\LARGE \bf
Do What You Say: Steering Vision-Language-Action Models via Runtime Reasoning-Action Alignment Verification

}

\author{%
  Yilin Wu$^{2*}$,
  Anqi Li$^{1}$,
  Tucker Hermans$^{1,3}$,
  Fabio Ramos$^{1,4}$,
  Andrea Bajcsy$^{2\S}$,
  Claudia P\'erez-D'Arpino$^{1\S}$%
  \thanks{\quad \quad \quad $^{1}$ NVIDIA.
  $^{2}$ Carnegie Mellon University.
  $^{3}$ University of Utah.
  $^{4}$ University of Sydney.
  $^{\S}$ Equal advising. 
  $^{*}$ Work done during an internship at NVIDIA.}%
}

\makeatletter
\let\@oldmaketitle\@maketitle%
\renewcommand{\@maketitle}{\@oldmaketitle%

\setcounter{figure}{0} %
    \centering
    \includegraphics[width=\linewidth]{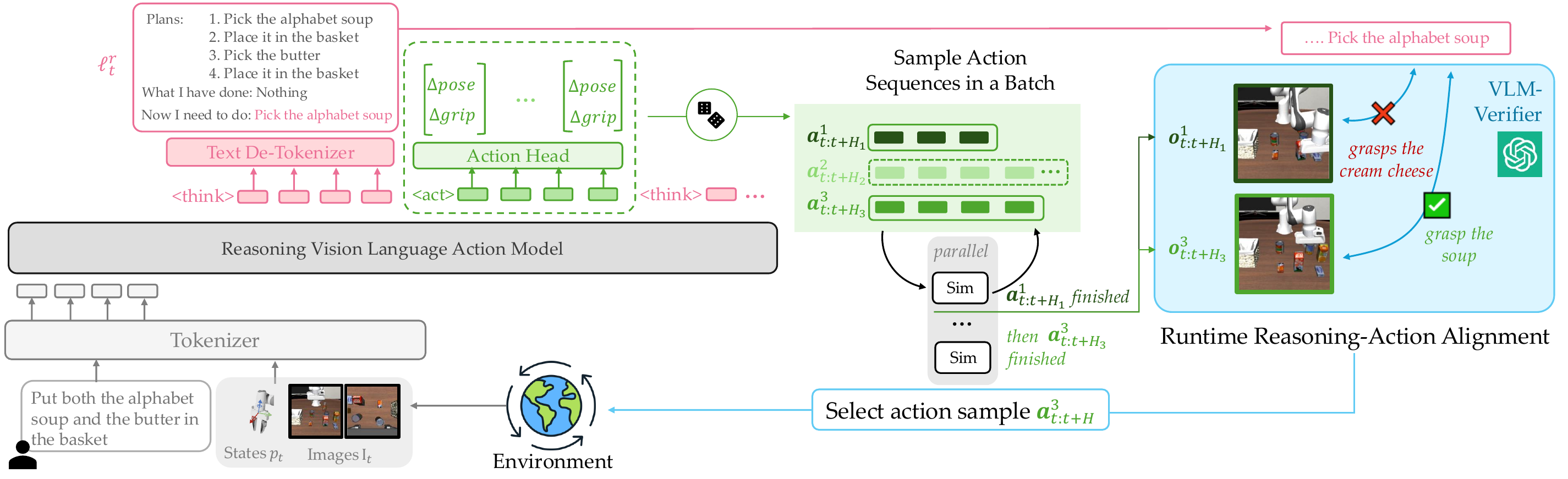}
    \captionof{figure}{\textbf{Method Overview.} Reasoning Vision Language Action (VLA) models interleave textual planning and action generation. After generating a text plan which describes intermediate goals, we sample a batch of action sequences, forward simulate their outcomes until the model switches to think again. We then use a Vision Language model (VLM) verifier to score alignment between the action's outcomes and the text plan. This improves the \textit{embodied CoT faithfulness} at runtime by executing only action samples that achieve the outcome of the text plan. 
    }
    \label{fig:method}
\vspace{-0.3in}
\bigskip}
\makeatother

\begin{document}

\maketitle

\thispagestyle{empty}
\pagestyle{empty}

\begin{abstract}
Reasoning Vision Language Action (VLA) models improve robotic instruction-following by generating step-by-step textual plans before low-level actions, an approach inspired by Chain-of-Thought (CoT) reasoning in language models. 
Yet even with a correct textual plan, the generated actions can still miss the intended outcomes in the plan, especially in out-of-distribution (OOD) scenarios.
We formalize this phenomenon as a lack of \emph{embodied CoT faithfulness}, and introduce a training-free, runtime policy steering method for reasoning-action alignment.
Given a reasoning VLA's intermediate textual plan, our framework samples multiple candidate action sequences from the same model, predicts their outcomes via simulation, and uses a pre-trained Vision-Language Model (VLM) to select the sequence whose outcome best aligns with the VLA's own textual plan. Only executing action sequences that align with the textual reasoning turns our base VLA's natural action diversity from a source of error into a strength, boosting robustness to semantic and visual OOD perturbations and enabling novel behavior composition without costly re-training. We also contribute a reasoning-annotated extension of LIBERO-100, environment variations tailored for
OOD evaluation, and demonstrate up to 15\% performance gain over prior work on behavior composition tasks and scales with compute and data diversity. 
Project Website at: \href{https://yilin-wu98.github.io/steering-reasoning-vla}{https://yilin-wu98.github.io/steering-reasoning-vla/}
\end{abstract}

\section{Introduction}

Vision Language Action (VLA) models~\cite{bjorck2025gr00t}\cite{black2024pi0}, which combine pre-trained vision-language backbones with action-generation heads fine-tuned on robot data~\cite{open_x_embodiment_rt_x_2023}, promise significant advances in generalizable robotic control. However, in practice, these models are not yet true generalists: their performance often degrades in novel scenes and over long horizons, and their ability to follow complex language instructions can be brittle~\cite{gao2025taxonomy}\cite{huang2025thinkact}\cite{kwok25robomonkey}.

To tackle these challenges and better exploit the vision-language backbone of the VLA, a growing body of work has introduced ``reasoning VLAs'' \cite{Zawalski24-ecot}\cite{lin2025onetwovlaunifiedvisionlanguageactionmodel}. This approach is inspired by the success of Chain-of-Thought (CoT) in large language models (LLMs): by producing intermediate step-by-step textual reasoning before generating a textual response, LLMs can significantly increase the complexity and reliability of language generation \cite{wei2022chain}.
Similarly, reasoning VLA’s implement a form of ``embodied CoT'' \cite{Zawalski24-ecot}: instead of generating actions directly, the model first generates intermediate textual reasoning, often grounded by visual observations, before producing reasoning-conditioned actions. For example, given the task description of ``Put both the alphabet soup and the butter in the basket'' in Fig.~\ref{fig:method}, the model generates the textual reasoning ``First I pick up the soup, then place it in the basket..., Now I need to pick up the soup'' and then generates a low-level action sequence (e.g., a sequence of relative end-effector poses). 

However, even with the perfect textual reasoning, the VLA-sampled action sequence may lead to different outcomes than the generated plan (e.g., grasping the cream cheese instead of the soup, as shown in Fig.~\ref{fig:method}). This  misalignment---where the outcomes of the generated actions fail to match the textual plan---is particularly prevalent in out-of-distribution (OOD) scenarios (novel instructions, objects, backgrounds, etc.), where enforcing alignment shows up to a 20\% performance gain in Sec.~\ref{sec:exp_ood}. This also mirrors observations from prior work on training reasoning VLAs~\cite{zhou2025chatvla}, where the reasoning capabilities of the models are stronger than language-conditioned action generation capabilities. 
Analogous to Chain-of-Thought (CoT) faithfulness in LLMs, which questions whether a model's generated reasoning accurately reflects the process used to derive an output~\cite{turpin2023language}, we formalize this reasoning-action misalignment as a robotics version of lack of \emph{embodied CoT faithfulness}: the expectation that outcomes described by a VLA's textual reasoning will be reliably realized by its subsequent low-level actions.

To address this gap, we introduce a \textbf{runtime reasoning-action alignment} method. This approach enforces embodied CoT faithfulness by actively steering a reasoning VLA's actions to match its own textual reasoning during execution. 
Specifically, right before the reasoning is updated, we autoregressively generate multiple action sequences from the base model, conditioned on the current textual plan and the observation from the preceding step (center, Fig.~\ref{fig:method}). 
We predict the outcomes of action samples via parallel simulation, and leverage an off-the-shelf Vision Language Model (VLM) verifier to score the alignment between the outcome and the preceding text plan. Only the action sequence that induces the correct outcome is executed on the robot.

Our contributions are summarized as follows:
\begin{itemize}
   
    \item \textbf{Runtime Policy Steering for Reasoning-Action Alignment:} we propose a new runtime framework for ensuring that a VLA's generated actions induce the outcomes described by its own textual reasoning. Even in in-distribution (ID) scenarios, we find that reasoning-action alignment improves task success by 8\% and preserves long-horizon semantic coherence without needing any additional fine-tuning data. More benefits are shown in OOD scenarios below.
    \item \textbf{Extensive Generalization Experiments and Scaling Analysis:} Through a controlled series of experiments on OOD shifts and behavior composition tasks in our extended version of LIBERO benchmark \cite{liu2023libero}, we find that our method outperforms relevant baselines by up to 15\%. This performance gap widens as the fine-tuning dataset grows, since a stronger policy generates a better set of candidate actions for our verifier to select from.
    \item \textbf{A Reasoning-Annotated VLA Dataset and an Extended Benchmark for Generalization Tests:} We contribute an open-sourced reasoning-annotated LIBERO-100 dataset for reasoning VLA training. We also contribute an extension of the LIBERO benchmark with new task descriptions, visual variations, and new behavior compositions. We use these for our evaluations but also plan to release it for the community to continue to study generalization capabilities of reasoning VLAs.

\end{itemize}

\section{Related Work}

\para{Vision Language Action Models \& Reasoning}
Large robotics datasets \cite{open_x_embodiment_rt_x_2023} have enabled the creation of generalist Vision Language Action (VLA) models \cite{bjorck2025gr00t}\cite{black2024pi0}. Despite their success, these VLA models often struggle with long-horizon tasks and out-of-distribution (OOD) generalization due to compounding errors and a limited ability to follow complex or abstract language instructions. To improve robustness, many approaches augment VLAs with Chain-of-Thought reasoning, either through hierarchical planner-controller dual systems \cite{huang2025thinkact} or an unified model that interleave intermediate reasoning and action generation \cite{Zawalski24-ecot}\cite{lin2025onetwovlaunifiedvisionlanguageactionmodel}\cite{intelligence2025pi}\cite{chen25training}. We build on the unified approach but identify a critical failure mode: a reasoning-action mismatch, where motor actions do not faithfully execute the model's self-generated textual plan.

\para{Runtime Optimization and Steering of Large Models}
Runtime optimization improves model's performance without costly retraining. In robotics, runtime steering has improved the alignment between human intent and robot actions in diffusion policies~\cite{wang2024inference}, and in VLA policies by sampling multiple action chunks and verifying them against learned critic. These critics are typically either offline Q-functions~\cite{nakamoto2024steering} (which can struggle in novel scenarios), or specially fine-tuned VLMs from synthetic preference labels which may not correlate with true task success 
\cite{kwok25robomonkey}. Critically, these methods verify actions in isolation and do not address the reasoning-action faithfulness gap—the challenge of ensuring that a sequence of low-level actions are semantically coherent with a model's self-generated textual plan. 
Inspired by faithfulness verification research in LLMs~\cite{turpin2023language}, our work explores a new direction: using runtime steering to explicitly enforce the alignment between a VLA's high-level reasoning and its low-level motor control.

\para{OOD Robustness \& Generalization}
A central motivation for improving reasoning-action alignment is to enhance out-of-distribution (OOD) robustness and generalization. To systematically evaluate such scenarios, the community has developed challenging benchmarks and OOD-related taxonomies \cite{gao2025taxonomy}\cite{liu2023libero}. However, existing OOD studies often focus on robustness to visual or semantic shifts (e.g., novel instructions, objects and backgrounds), with less attention paid to compositional generalization: the ability to recombine learned skills to solve new tasks. We adopt the LIBERO benchmark~\cite{liu2023libero} for its focus on long-horizon, compositional tasks. However, as it lacks the clear separation of different OOD factors, we extend it to systematically study OOD robustness and generalization. 
Furthermore, while recent work has studied OOD failure detection \cite{agiaunpacking} and mitigation via observation interventions \cite{ gupta2025adapting}, our method focuses on improving compositional generalization via runtime verification, enabling the policy to solve unseen tasks by executing actions that achieve its own plan.

\section{Problem Formulation}
\label{sec:problem_formulation}

We contextualize our problem setting by first contrasting ``vanilla'' VLAs with reasoning VLAs. We then  formally define the \textit{embodied CoT faithfulness gap} between a reasoning VLA's plans and the outcome of its actions that we address in Sec.~\ref{sec:approach}.

\para{Setup and Notation}
We consider general robotic manipulation settings. Let $\lang^g$ denote the high-level natural language instruction for a task (e.g., clean the table). At each real timestep $t$, the robot's observation is $o_t = (I_t, p_t)$, where $I_t$ is the visual observation including, e.g., wrist and agent-view RGB images, and $p_t \in \mathbb{R}^d$ is the proprioceptive state of the robot. The robot's policy maps the current observation $o_t$ and language instruction $\langg$ to action $a_t$. 

\para{Vanilla VLA Formulation}
Traditional vision-language-action (VLA) models~\cite{black2024pi0} learn policies end-to-end via a demonstration dataset of $N$ trajectories, $\mathcal{D} = \{(\traj, \langg)_i\}_{i=1}^N$, where each trajectory $\traj$ consists of $T$ observation-action pairs, 
$\traj = \{(o_{t}, a_{t})\}_{t=1}^{T}$. 
The VLA policy $\pi_\theta^{\text{vla}}$ maps the observation and instruction $(o_t, \lang^g)$ to a distribution over actions $a_t$. 
The training objective is to  maximize the log-likelihood of the policy under the expert data distribution:
\begin{equation}
    \mathcal{L}_{\text{vla}}(\theta; \mathcal{D}) = \sum_{(o_t, a_t, \langg)\in \mathcal{D}}
    -\log \pi_\theta^{\text{vla}}(a_t \mid o_t, \lang^g).
\end{equation}
\vspace{-0.4cm}

During inference, at any timestep $t$, the policy samples an action to execute given the current observation and language instruction: $\hat{a}_t \sim \pi_\theta^{\text{vla}}(\cdot  \mid o_t, \lang^g)$.
However, vanilla VLA models often degrade on long-horizon tasks~\cite{huang2025thinkact}, as errors compound over extended trajectories and complex language instructions are difficult to decompose and ground in actions.

\para{Reasoning VLA Formulation}
To mitigate the challenges of long-horizon planning, reasoning VLAs \cite{Zawalski24-ecot}\cite{lin2025onetwovlaunifiedvisionlanguageactionmodel}\cite{ intelligence2025pi} factor the problem by adding the generation of Chain-of-Thought (CoT) in the form of intermediate reasoning. To learn this ``embodied'' version of CoT, reasoning VLAs are trained on a reasoning-annotated demonstration dataset with fine-grained textual plans. Let this dataset be denoted by $\mathcal{D}_{\text{reason}}= \{(\traj^{\text{reason}}, \langg)_i\}_{i=1}^N$, where each trajectory $\traj^{\text{reason}}_i$ is divided into $L_i$ segments: $\traj^{\text{reason}}_i = \{(\oseq_j, \aseq_j, \langropt_j)\}_{j=1}^{L_i}$.Here, $\langr_j$ is the text plan for the $j$-th segment, with corresponding observation and action sequences $\oseq_j = \obs_{t_j:t_j+H_j}$ and $\aseq_j=a_{t_j:t_j+H_j}$ of varied lengths. For notational simplicity, we drop the subscript $j$ from the start time $t$ and use $\oseq_j = \obs_{t:t+H}$ and $\aseq_j=a_{t:t+H}$ hereafter.

Given the reasoning dataset $\mathcal{D}_{\text{reason}}$, a reasoning VLA $\pi_\theta^{\text{r-VLA}}$ is trained via supervised fine-tuning (SFT) to generate both (i) a textual plan as intermediate reasoning and (ii) the subsequent plan-conditioned action sequence \cite{lin2025onetwovlaunifiedvisionlanguageactionmodel}\cite{intelligence2025pi}. Specifically, $\pi_\theta^{\text{r-VLA}}$ takes as input the tokenized observations, language instruction and intermediate text plans, and generates tokens that are decoded either as text or actions. The training objective is a weighted sum of the text loss and action loss:
\begin{equation}
\label{eq:rvla_loss}
\begin{aligned}
\mathcal{L}_{\text{r-vla}}(\theta; \mathcal{D}_{\text{reason}}) &= \sum_{\mathclap{{(o_{t:t+H},a_{t:t+H}, \langropt_j, \langropt_{j-1}, \langg) \in \mathcal{D}_{\text{reason}}}}}\lambda_{\text{reason}}\,
\mathcal{L}_{\text{reason}}+\lambda_{\text{act}}\,\mathcal{L}_{\text{act}}   \\
\mathcal{L}_{\text{reason}} &=  -\log \pi_\theta^{\text{r-vla}}(\langropt_j \mid o_t, \langropt_{j-1}, \langg) \\
\mathcal{L}_{\text{act}} &=-\sum_{t'=t}^{t+H} \log \pi_\theta^{\text{r-vla}}(a_{t'} \mid o_{t'},\langr_j, \langg )
\end{aligned}
\end{equation}
where $ 0 < \lambda_{\text{reason}} < 1$ and $0< \lambda_{\text{act}} < 1$ and $\lambda_{\text{reason}}+ \lambda_{\text{act}}=1 $.

During inference, at real timestep $t$, given the text plan $\langrhat_{\text{last}}$ from last reasoning step, the policy can either update the current textual plan (e.g, ``pick up the soup can'') via $\langrhat \sim \pi_\theta^{\text{r-vla}}(\cdot \mid o_t, \langrhat_{\text{last}}, \lang^g)$  or generate the action conditioned on the last textual reasoning: $\hat{a}_t \sim \pi_\theta^{\text{r-vla}}(\cdot \mid o_t, \langrhat_{\text{last}}, \lang^g)$.
This decomposition aims to leverage the strong reasoning and language understanding capabilities of the pre-trained VLM backbone. In Sec.~\ref{sec:method_vla}, we discuss the specific reasoning VLA~\cite{lin2025onetwovlaunifiedvisionlanguageactionmodel} we build our method on.

\para{Embodied Chain-of-Thought (CoT) Faithfulness Gap}
\label{sec:problem_ecot_gap}
In factorized formulations of reasoning VLAs, textual reasoning and action generation are optimized together. Prior work~\cite{zhou2025chatvla} has shown that while the pre-trained VLM backbone of a VLA can be efficiently finetuned for textual planning with sparse data, the same is not necessarily true for low-level control. Our experiments confirm that acquiring robust control policies conditioned on the text plan is a far more challenging problem: for example, the text generation is largely reliable (100\% accuracy for in-distribution tasks trained with LIBERO-10-R dataset in Sec.~\ref{sec:exp_alignment}), while most task failures stem from the action generation module failing to execute that plan.

Given that the textual plan is often correct or can be easily corrected by an external agent~\cite{Zawalski24-ecot}, the critical bottleneck is the policy's ability to reliably follow its own reasoning. We define this core challenge as the \emph{Embodied Chain-of-Thought Faithfulness Gap}: the misalignment between a generated textual plan and the physical outcome of the associated low-level actions. 

\para{Goal} In this work, our goal is to minimize this gap at each timestep $t$, which can be formalized as minimizing the misalignment loss for candidate action sequence $\hat{a}_{t:t+H}$ sampled from $\pi_\theta^{\text{r-vla}}$:
\begin{equation}
    \mathcal{L}_\text{align}(\theta; o_t, \langrhat, \langg) = -\mathbb{E}_{\mathcal{P},\pi_\theta^\text{r-vla}} \left[ R_{\text{align}}({o}_{t:t+H}, \langrhat) \right],
    \label{eq:reward-plan-alt}
\end{equation}
where the expectation is taken over the stochastic outcomes of the environment's true dynamics $\mathcal{P}$ and the reasoning VLA model, i.e., ${o}_{t'+1} \sim \mathcal{P}(\cdot | {o}_{t'}, \hat{a}_{t'})$, $\hat{a}_{t'} \sim \pi_\theta^{\text{r-vla}}(\cdot | {o}_{t'}, \langrhat, \langg)$, for all $t'\in [t,t+H]$ and 
Here, $R_{\text{align}}(o_{t:t+H}, \langrhat)$ is a reward function that measures whether the future observation $o_{t:t+H}$ satisfies the subgoal described by the textual plan $\langrhat$. %

Crucially, this alignment objective is distinct from the standard behavior cloning loss (e.g., Eq.~\ref{eq:rvla_loss}) used to train the reasoning VLA. Since the model is not explicitly optimized to maximize this alignment, the actions it generates can be imprecise at best or pursue an entirely different subgoal than the one articulated in the textual plan at worst.

\section{Approach: Steering for Embodied Reasoning-Action Alignment (\acronym)}
\label{sec:approach}
Our key idea is to enforce \emph{Embodied CoT Faithfulness} at runtime, thereby ensuring that actions produced by a reasoning VLA model semantically realize the model’s own intermediate textual plans. 
To achieve this, we first train a reasoning VLA model from prior open-source work~\cite{lin2025onetwovlaunifiedvisionlanguageactionmodel} and then propose a runtime policy steering procedure to select the actions that maximize reasoning-action alignment. We call our overall steering approach \textbf{\acronym}: \textbf{S}teering for \textbf{E}mbodied reasoning-action \textbf{AL}ignment.

\subsection{Training a Reasoning VLA Model}

Throughout our experiments, we train reasoning VLAs in a controlled way to study the capabilities conditioned on a specific fine-tuning dataset. 
To enable this controlled fine-tuning, we first propose a pipeline to automatically annotate any given demonstration dataset with intermediate textual reasoning. Using this annotated dataset, we then fine-tune the $\pi_0$ base model~\cite{black2024pi0}, following the training recipe from~\cite{lin2025onetwovlaunifiedvisionlanguageactionmodel}, resulting in a VLA capable of interleaving text reasoning with action generation.

\para{Reasoning Data Annotation Pipeline} 
\label{sec:data_annotation}
Most robot demonstration datasets contain only a single high-level language text instruction $\langg$ per episode. 
To obtain intermediate reasoning labels, we introduce an automated annotation pipeline that decomposes long-horizon demonstrations into intermediate text plans. 
As shown in Fig.~\ref{fig:method}, we modify the reasoning format from prior work~\cite{lin2025onetwovlaunifiedvisionlanguageactionmodel} and define an intermediate reasoning step $\langr$ to abide by the following format:
\begin{itemize}
    \item \textit{Plans}: \textless all text plans in the task\textgreater
    \item \textit{What has been done}: \textless completed plans\textgreater
    \item \textit{Now I need to do}: \textless the next text plan to execute\textgreater
\end{itemize}

To avoid the expensive manual annotation of sub-task boundaries required by prior work~\cite{lin2025onetwovlaunifiedvisionlanguageactionmodel}, we use Gemini~\cite{comanici2025gemini} to automatically generate annotations given the input of demonstration videos and their high-level task instructions ($\langg$). In a single pass, Gemini jointly generates a one-sentence textual plan ($\langrhat_j$) and its corresponding end-timestep $t'_j$ for each sub-task. Since one sub-task's end marks the next one's start, this process directly yields a sequence of annotations: $(\langrhat_1, 1), (\langrhat_2, t'_1), \dots, (\langrhat_L, t'_{L-1})$. The number of sub-tasks L varies per episode based on Gemini's decomposition, and we manually verify all generated annotations for quality (details in Appendix~\ref{appendix:annotation} on our website).

\para{Reasoning VLA Training}
\label{sec:method_vla}
While our proposed runtime steering method is model-agnostic, we instantiate our system using a reasoning VLA based on recent   work~\cite{lin2025onetwovlaunifiedvisionlanguageactionmodel}. This architecture features an adaptive switching capability between text and action generation. 
The core of this mechanism lies in two special tokens, \texttt{<think>} and \texttt{<act>}, which function as switching signals. When the model generates \texttt{<think>}, subsequent tokens are decoded as textual reasoning until an end-of-sentence token is reached. When the model generates \texttt{<act>}, the following tokens are passed to an action head, which produces a continuous action sequence through a diffusion process in Fig.~\ref{fig:method}. This adaptive switching allows the VLA model to generate variable-length action sequences tailored to the complexity of the its own textual plan, effectively segmenting tasks into a series of text-plan-conditioned action sequences.

For training, we follow the recipe from~\cite{lin2025onetwovlaunifiedvisionlanguageactionmodel}, fine-tuning the pre-trained $\pi_0$ VLA on our reasoning-annotated dataset. 
The training data is formatted by  prepending the \texttt{<think>} token to each textual reasoning label 
$\langr$ and prepending the \texttt{<act>} token to its corresponding action sequence. The model is then fine-tuned end-to-end with the combined loss function in Eq.~\ref{eq:rvla_loss}, which uses a cross-entropy loss for text generation and a flow-matching loss for action generation.

\subsection{Runtime Policy Steering via Semantic Verification}

As formalized in Sec.~\ref{sec:problem_ecot_gap}, ensuring that the action sequence will cause outcomes that match with the text plan requires solving the alignment objective in Eq.~\ref{eq:reward-plan-alt}. This presents two fundamental challenges. 
\begin{itemize}
    \item \textbf{Direct Optimization}: The loss $\mathcal{L}_{\text{align}}$
  is hard to optimize directly via backpropagation due to the non-differentiable dynamics $\mathcal{P}$ and action sampling from $\pi_\theta^{\text{r-vla}}$. 
 \item \textbf{Reward Modeling}: The true alignment reward $R_{\text{align}}$
  is hard to model analytically or via heuristics, as it requires complex and contextual semantic judgment. 
\end{itemize}

To overcome these obstacles, we introduce a novel runtime policy steering method that provides a tractable, approximation of the reasoning-action alignment problem. Our approach circumvents the direct optimization challenge using a best-of-N sampling strategy, an effective technique in LLMs~\cite{snell2024scaling}. 
We address the challenge of reward modeling by using the strong commonsense and physical reasoning capabilities of pre-trained VLMs as our open-world verifier \cite{agiaunpacking}\cite{wu2025forewarn}\cite{gao2024physically}.
Our framework is a three-stage process, which we detail next.

\para{Runtime Policy Steering: Hypothesize, Predict, and Verify} 
Our runtime steering approach unfolds in a three-stage process at each reasoning step, after a new textual plan $\langrhat_t$ is generated.

\para{1. Hypothesize}
In this stage, the policy generates multiple plausible sequences of actions. We sample a set of $K$ candidate action sequences in parallel from the policy $\pi_\theta^{\text{r-vla}}$, conditioned on $(o_t, \langrhat, \langg)$. 
The generation of each sequence is autoregressive, meaning that each action is sampled one step at a time. The process terminates when the model outputs a $\texttt{<think>}$ token, resulting in variable-length action sequences (shown in center, Fig.~\ref{fig:method}). 
This process yields a set of hypotheses:
\begin{equation}
A_t = \{ \aseqhat_t^{(k)} \sim \pi_\theta^{\text{r-vla}}(\cdot \mid o_t, \langrhat, \langg) \}_{k=1}^K
\end{equation}
where each $\aseqhat_t^{(k)}=a^{(k)}_{t:t+H_k}$ has a different horizon $H_k$\footnote{The sequence $\aseqhat_t^{(k)}$ is generated autoregressively from the model, rather than in a single forward pass. }.

\para{2. Predict}
The \emph{Predict} and \emph{Hypothesize} stages are tightly coupled in an interleaved loop. This is necessary because the policy is autoregressive: generating the action $\hat{a}_{t'}$ requires the predicted observation $\hat{o}_{t'}$.
We use a dynamics model, $\hat{\mathcal{P}}$, that approximates the ground-truth dynamics $\mathcal{P}$ to generate the observation sequence $\hat{\mathbf{o}}^{(k)}_t$ given $\aseqhat_t^{(k)}$. In simulation, we use parallel environment instances for high-fidelity predictions ~\cite{liu2023libero}; in the real world, a learned world model~\cite{alhaija2025cosmos}  or digital twin~\cite{mu2025robotwin} would serve this role. This \emph{Hypothesize-Predict} loop continues until the next think token is generated.

\para{3. Verify} In the final stage, we select the most aligned action sequence by leveraging a pre-trained VLM (e.g., GPT-4o) as a proxy $R_\psi$ for the ground-truth alignment reward, $R_{\text{align}}$. The VLM verifier assigns a score to each candidate sequence by evaluating its predicted outcome against the textual plan.

The action sequence that receives the highest alignment score from the VLM is chosen for execution.
The VLM is prompted to return a binary score $R_\psi \in \{0,1\}$ indicating whether the predicted outcome successfully fulfills the plan. For a practical speedup, we only input the initial image of the episode $I_1$, the predicted final image from the candidate sequence $\hat{I}^{(k)}_{t+H_k}$, and the textual plan ($\langrhat$) to the VLM. More details are in Appendix~\ref{appendix:steering} on our website.

\para{Runtime Considerations}
Our framework is optimized for low latency using parallel and asynchronous execution. The $K$ action sequences are generated in parallel via a tightly coupled \emph{Hypothesize-Predict} loop. Crucially, the \emph{Verify} stage runs asynchronously, immediately evaluating each candidate as it completes generation. Since the sequences have variable lengths and finish at different times, this enables an early-exit strategy: we execute the \textbf{first} sequence that the VLM successfully verifies, rather than waiting to evaluate all $K$ candidates. 
This significantly reduces decision-making latency (see Appendix~\ref{appendix:runtime} on our website).

\begin{figure*}[!htbp] %
  \centering

  \begin{subfigure}{0.65\linewidth}
    \includegraphics[width=\linewidth]{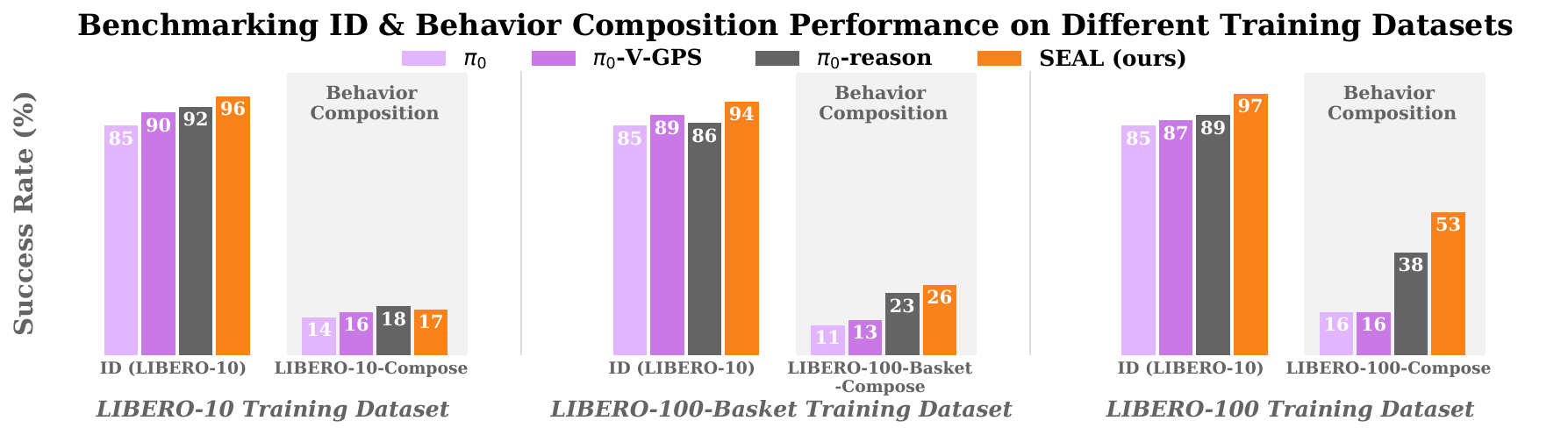}
    \caption{\textbf{In-distribution \& Compositional Generalization Success Rates.} 
Each model is evaluated by averaging the performance of all tasks in its corresponding suite in Sec~\ref{sec:task_suites}, with 50 trials per task. 
}
    \label{fig:exp_compose}
  \end{subfigure}\hfill
  \begin{subfigure}{0.3\linewidth}
    \includegraphics[width=\linewidth]{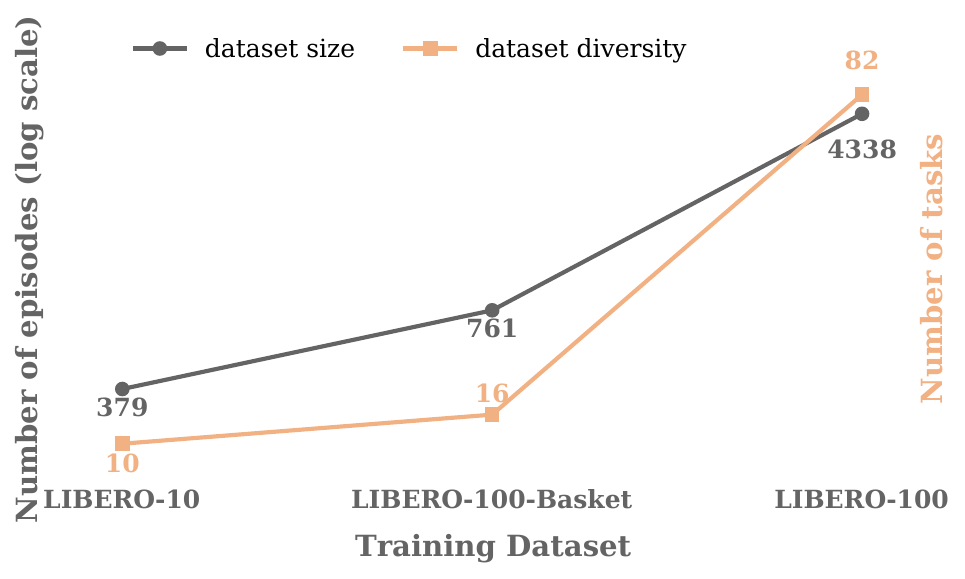}
    \caption{\textbf{Dataset Statistics.} Comparison of diversity (number of tasks) and sizes (number of episodes) of our three datasets.}
    \label{fig:data_stats}
  \end{subfigure}

  \caption{\textbf{Dataset Scaling Results.} Our method, \textbf{\ours}, consistently outperforms all baselines by enforcing reasoning-action alignment with increased performance as the training data scales, especially on challenging behavior composition tasks. }
  \label{fig:twopanels}
  \vspace{-0.4cm}
\end{figure*}

\section{Experiments}
We conduct a series of experiments in a simulation benchmark~\ref{sec:exp_benchmark} to carefully study: (1) the benefit of enforcing reasoning-action alignment at runtime (Sec.~~\ref{sec:exp_alignment}), (2) robustness to out-of-distribution (OOD) shifts and generalization to new behavior compositions (Sec.~\ref{sec:exp_ood}), and (3) performance and runtime scaling as a function of the number of samples to be verified (Sec.~\ref{sec:exp_scaling}). Videos can be found on our website. 

\subsection{Benchmark \& Implementation Details}
\label{sec:exp_benchmark}

\para{Training Dataset}
\label{sec:datasets}
We use the LIBERO benchmark \cite{liu2023libero} due to its diverse, long-horizon manipulation tasks.
To investigate how performance scales with the size and diversity of the VLA fine-tuning data, we use three datasets of increasing scale. Each dataset is processed with the reasoning annotation pipeline described in Section~\ref{sec:data_annotation}. \textbf{\textit{Libero-10-R}}: Our smallest reasoning dataset, consisting of demonstration data for the \textit{LIBERO-10} evaluation tasks.
\textbf{\textit{Libero-100-Basket-R}}: A medium-sized dataset we create to test skill composition, supplementing \textit{Libero-10} with a curated subset of tasks from \textit{Libero-90} that involve ``pick and place into basket'' skills.
\textbf{\textit{Libero-100-R}}: Our largest dataset, generated from annotating all demonstrations from the full \textit{LIBERO-100} suite.

\para{Evaluation}
\label{sec:task_suites}
We evaluated our method in three suites of tasks, each of which involved running 50 trials.
\textit{\textbf{(1) In-Distribution (ID) Performance}} (Sec.~\ref{sec:exp_alignment}) uses the test set from the \textit{LIBERO-10} benchmark which has 10 long-horizon manipulation tasks.
\textit{\textbf{(2) Robustness to OOD Shifts}} (Sec.~\ref{sec:exp_ood}) varies the visual and language instruction based on the taxonomy proposed by~\cite{gao2025taxonomy}. We construct four OOD variants of the \textit{LIBERO-10} suite, each of which contains 10 tasks from the base suite. Among them, two variants are modified with   \textit{semantic OOD} changes where
instructions are rephrased but the object descriptions are the same (\textbf{LIBERO-10-Lang-Rephrase}),
or object descriptions are changed using alternate properties (\textbf{LIBERO-10-Lang-Object-Property}). We also have \textit{visual OOD} changes, where objects are added or non-target objects are replaced (\textbf{LIBERO-10-Visual-Scene}), or the background and camera viewpoints are changed (\textbf{LIBERO-10-Visual-Viewpoint}).

Finally, \textit{\textbf{(3) Generalization to Behavior Composition}} (Sec.~\ref{sec:exp_alignment})
is tested in a suite of new long-horizon tasks that recombine learned skills, while sharing the same initial states as the original training demonstrations. For example, a model trained on tasks like ``put A and B in the basket'' and ``put C in the basket'' is then evaluated on the unseen instruction ``put A and C in the basket.'' 
\textbf{LIBERO-10-Compose} has two new compositional basket-related pick-and-place tasks, 
\textbf{LIBERO-100-Basket-Compose} has nine novel compositional basket-related tasks, 
and \textbf{LIBERO-100-Compose} has thirteen novel compositional tasks across a wider range of objects and tasks. The full list of these tasks and their videos are on our website.

\para{Baselines}
We compare our approach, \ours, to three baselines all based on the same $\pi_0$~\cite{black2024pi0} architecture. \textbf{\vla} is a vanilla VLA~\cite{black2024pi0} that directly maps visual-language inputs to actions without any intermediate textual reasoning. \textbf{\reasonvla} is the base reasoning VLA model~\cite{lin2025onetwovlaunifiedvisionlanguageactionmodel} trained with our reasoning-annotated data without runtime verification and steering. It is a direct ablation of our primary contribution.
\textbf{\vgps}~\cite{nakamoto2024steering} is a state-of-the-art runtime steering method\footnote{Another recent runtime steering method is~\cite{kwok25robomonkey}, but its pre-trained action verifier for LIBERO has not been released, precluding a fair comparison.}. It trains a critic Q-function using offline reinforcement learning~\cite{kostrikovoffline} and, at runtime, executes the action chunk with the highest Q-value from a set of policy samples. 
In contrast, \ours verifies an entire action sequence against a textual plan using a pre-trained VLM as the critic.
For \vgps and \ours, we use $K=10$ candidate samples per reasoning step unless otherwise specified. We choose the temperature of 1 for \vgps as it has the best average performance.

\para{Training} Our baseline reasoning VLA model, \textbf{\reasonvla}, is created by fully fine-tuning the 
$\pi_0$ \cite{black2024pi0} backbone on our reasoning-annotated dataset with $\lambda_{\text{reason}} =0.5 , \lambda_{\text{act}} =0.5 $ as loss coefficients. Training \textbf{\reasonvla} takes approximately 20 hours on 8 A100 GPUs, whereas finetuning \vla takes about 6 hours. More details are in Appendix.~\ref{appendix:training}

\subsection{On The Value of Verifying Reasoning-Action Alignment}
\label{sec:exp_alignment}

We first study the value of verifying reasoning-action alignment by comparing \ours against the three  baselines from above
on both in-distribution tasks and the challenge of novel behavior composition tasks. 

   \begin{figure*}
        \centering
        \includegraphics[width=\linewidth]{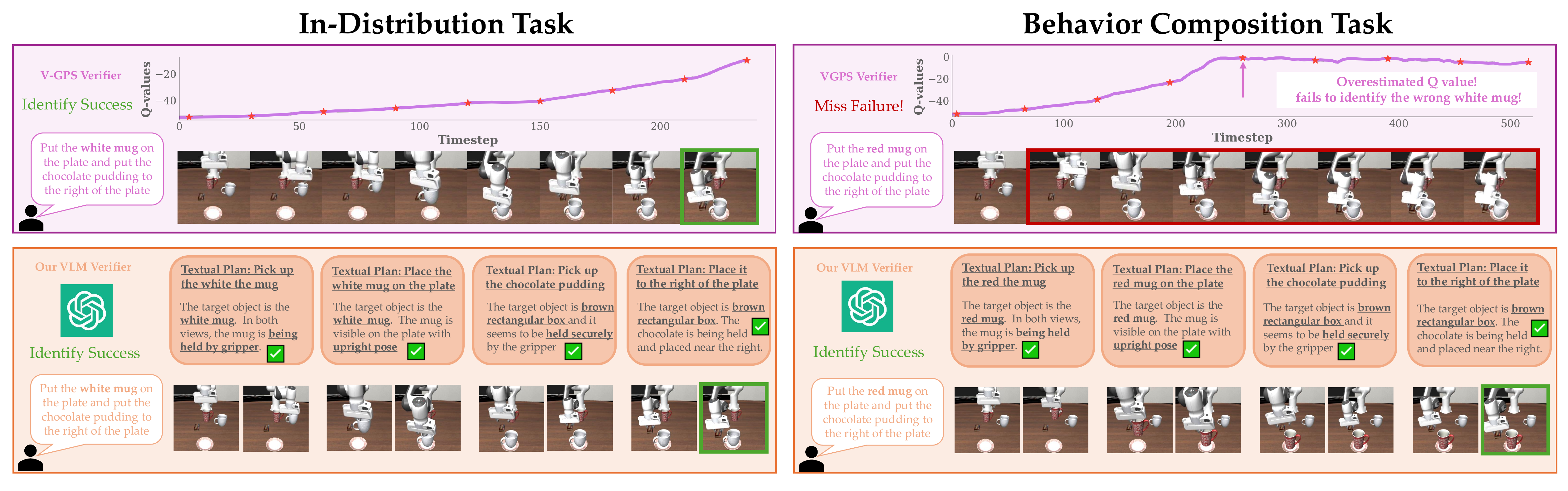}
        \caption{\textbf{Comparison Between Verification Methods.}  (left) For ID tasks, both the Q-value based verifier in \vgps and our VLM verifier can accurately predict task progress.  (right) For behavior composition tasks requiring stronger generalization capabilities, the Q-value overestimates task progress, failing to provide useful signal to guide \vla. However, our VLM verifier preserves the commonsense reasoning and correctly guides the base \textbf{\reasonvla} policy towards the task goal.}
        \label{fig:q-value}
        \vspace{-0.6cm}
    \end{figure*}

\para{Results: In-Distribution (ID) Tasks}
On ID tasks from \textit{LIBERO-10}, Fig.~\ref{fig:exp_compose} (left panel for each dataset), our method \textbf{\ours} achieves the highest success rates of 94-97\%, and both the base reasoning VLA \textbf{\reasonvla} and \vgps outperform the no-reasoning \vla~policy. 

Since these task instructions match the training distribution of \vgps, this method can provide reliable guidance, with the steady increase in Q-values shown in Fig.~\ref{fig:q-value} (left). 
The place where \textbf{\ours} shines is addressing the inherent limitations of chunk-based steering. Since the Q-function evaluates short fixed-length action chunks, it can oscillate between choices between timesteps. In contrast, our method verifies at the \textit{plan-level}---where the variable-length action sequences correspond to the model's textual plan---filtering out executions that are imprecise or correspond to the wrong subgoal, thus achieving a higher success rate.

\para{Results: Novel Behavior Composition}
To test the ability to reuse learned low-level behaviors for novel instructions, we evaluate on behavior composition tasks from Sec.~\ref{sec:exp_benchmark}.

In Fig.~\ref{fig:exp_compose} (right panel for each dataset), we see 
that as the VLA training dataset grows, the reasoning-enabled models (\textbf{\reasonvla} base and \textbf{\ours} verified) show substantial performance gains while \vla and \vgps stagnate at a low success rate ($\sim$15\%). 
We hypothesize that this is because the weak language understanding of \vgps (MUSE encoder~\cite{yang2020multilingual}) fails to generalize to novel instructions and gives misleading value signals, shown in Fig.~\ref{fig:q-value} (right).

In contrast, the reasoning in \textbf{\reasonvla} exhibits stronger generalization. We hypothesize this is due to the strong VLM backbone that can generate correct plans that decompose novel long-horizon problems into familiar subtasks. However, its performance is ultimately capped by a new challenge: as the training data and skill diversity grow, the policy's action generation becomes more varied, increasing the likelihood of misalignment with its own correct plan.

Our method, \ours, is designed specifically to resolve this tension. By verifying and selecting the action sequence that faithfully executes the generated plan, it leverages this action diversity as a strength rather than a source of errors. 
Overall, \ours 
exhibits a positive scaling trend, especially on large and diverse training datasets like \emph{LIBERO-100-R}.

\subsection{On the Value of Verification for Robustness to OOD shifts}
\label{sec:exp_ood}

We further evaluate the OOD robustness of each method on four challenging distribution shifts, from Sec.~\ref{sec:task_suites}: two semantic shifts (\textit{Lang-Rephrase}, \textit{Lang-Object-Property}) and two visual shifts (\textit{Visual-Scene}, \textit{Visual-Viewpoint}).

\begin{figure}[!h]
    \centering
    \includegraphics[width=\linewidth]{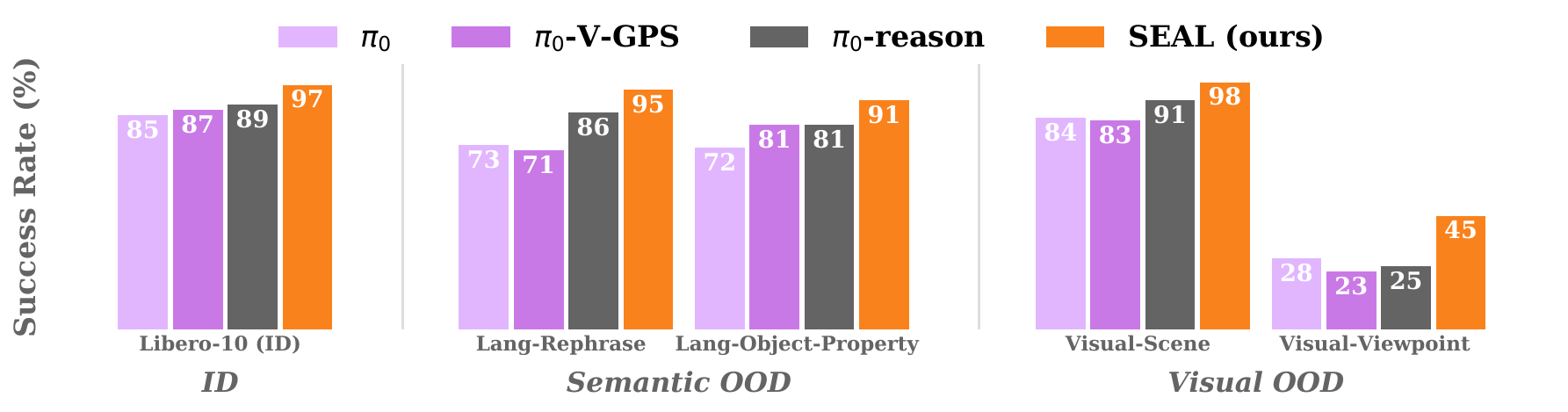}
    \caption{\textbf{Robustness to Semantic and Visual OOD Shifts.} We evaluate four different OOD variations for each task in \textit{Libero-10}. The performance is averaged across 10 tasks in the task suite with 50 trials per task. \ours maintains the best performance across all OOD shifts, even in the most challenging Visual-Viewpoint OOD scenario. }
    \label{fig:ood-generalization}
    \vspace{-0.6cm}
\end{figure}

\para{Results: Semantic OOD Robustness}
We find that reasoning-based models are most robust (Fig.~\ref{fig:ood-generalization}), while non-reasoning baselines like \vla and \vgps prove brittle. For instance, the base \vla policy’s performance drops by 12\% on rephrased instructions alone, while \vgps's success rate falls as low as 71\%. In contrast, generating intermediate textual plans makes \textbf{\reasonvla} highly robust to paraphrasing.
However, we found that spurious word-object correlations can still lead to action errors for \textbf{\reasonvla}. Our method, \ours, inherits this linguistic robustness but further improves upon it by using the verification stage to filter out the resulting action errors, consistently maintaining a success rate above 91\%.

\para{Results: Visual OOD Robustness}
We find that replacing distractor objects (\textit{Visual-Scene}) minimally influences most methods, but causes a notable performance drop (4\%) for \vgps. We hypothesize that its Q-function has overfit to the specific visual context of the training demonstrations.

In contrast, changing the viewpoint and background (\textit{Visual-Viewpoint}) is the most challenging test, causing performance to drop significantly across all methods. We hypothesize that this is influenced by the lack of viewpoint and background diversity in the training data for the same task. However, here we see the most pronounced benefit of our method: by filtering out noisy actions caused by the severe visual shift, \textbf{\ours} maintains a 45\% success rate, outperforming the next-best baseline by over 17\%.

\begin{figure}[!ht]
    \centering
    \includegraphics[width=\linewidth]{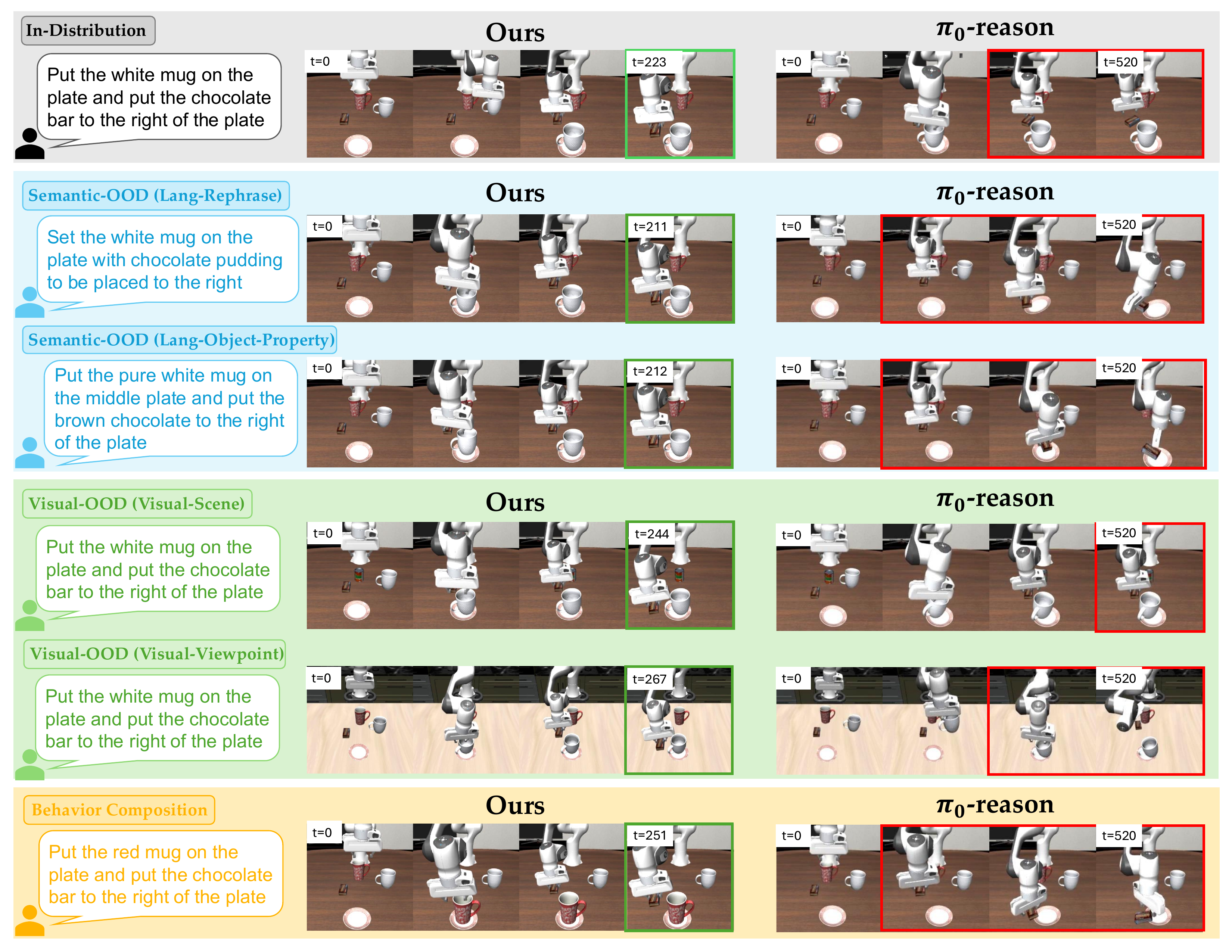}
    \caption{\textbf{OOD Shifts \& Behavior Composition Examples.} We visualize \ours and \textbf{\reasonvla}. We mark a successful rollout with a green box and a failure with red. \ours (on the left) can reliably generate successful rollouts to complete the tasks. \textbf{\reasonvla} policy (on the right) fails due to imprecision or by completing a totally different textual plan. }
    \label{fig:ood_qualitative}
    \vspace{-0.6cm}
\end{figure}

\para{Summary}
Across four stress-tests, \textbf{\ours} is consistently the most robust method, maintaining a strong quantitative and qualitative performance improvement (shown in Fig.~\ref{fig:ood_qualitative}). Our results also suggest that current VLA models are far more vulnerable to major visual shifts (like viewpoint and background changes) than to semantic or minor scene changes.


\begin{figure*}[!ht] %
  \centering

  \begin{subfigure}{0.32\linewidth}
    \includegraphics[width=\linewidth]{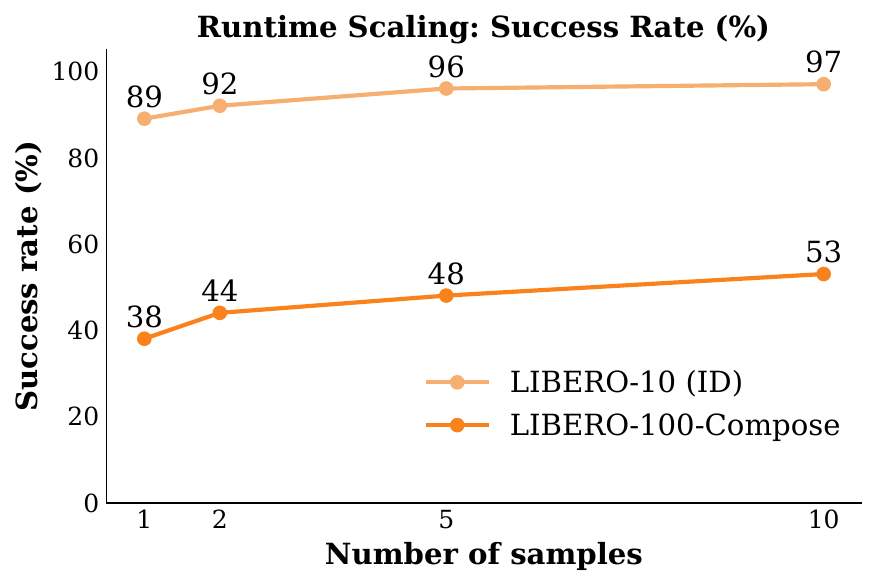}
    
  \end{subfigure}\hfill
  \begin{subfigure}{0.32\linewidth}
    \includegraphics[width=\linewidth]{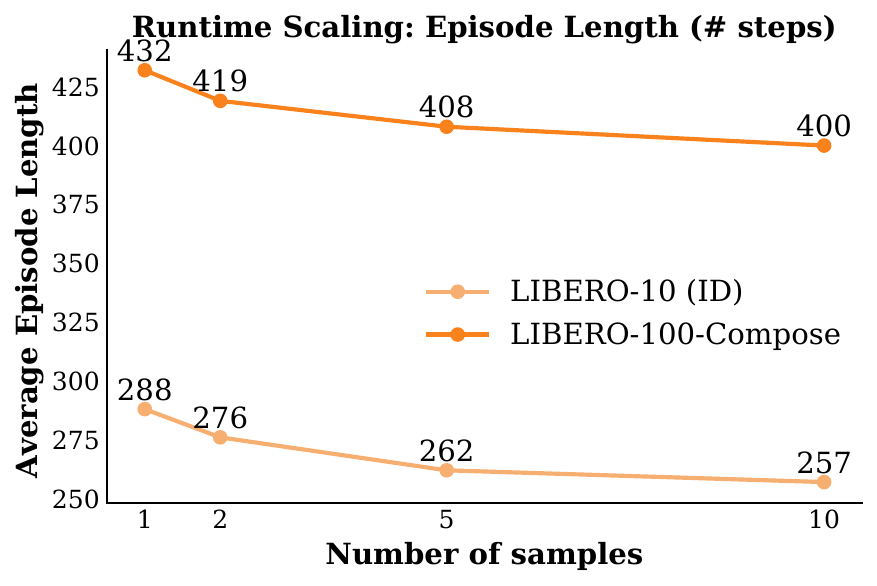}

  \end{subfigure}
\begin{subfigure}{0.32\linewidth}
    \includegraphics[width=\linewidth]{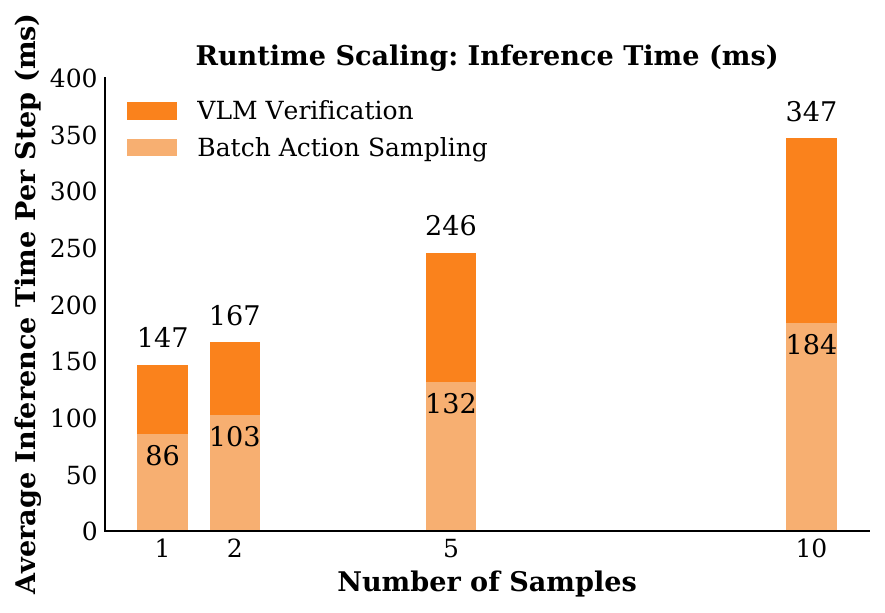}

  \end{subfigure}
      \caption{\textbf{Runtime Scaling Law.} Increasing the number of candidate sequences ($K$) boosts success rates (\textbf{Left}) on in-distribution (ID) and behavior composition tasks, while reducing steps for task completion. (\textbf{Middle}). The inference time (\textbf{Right}) scales favorably, with our latency at $K=10$ (347ms) lower than the $\sim$520ms in~\cite{kwok25robomonkey}. Results are averaged over 50 trials per task.}
  \vspace{-0.6cm}

  \label{fig:test_time_scaling}
\end{figure*}
\subsection{Runtime Scaling Law}
\label{sec:exp_scaling}
We find that increasing the number of candidate action sequences ($K$) at runtime directly boosts performance. As $K$ increases from one to ten, \ours enjoys higher success rates--with a notable 15\% gain on challenging compositional tasks---and shorter episode lengths (Fig.~\ref{fig:test_time_scaling}).
Moreover, our method’s inference time scales linearly with the number of samples and grows more slowly than that of~\cite{kwok25robomonkey}. The main bottleneck is the VLM verifier query time. This linear trend indicates that we can obtain a practical performance boost at runtime simply by increasing the number of candidates for verification.

However, the performance gains exhibit diminishing returns as the performance is ultimately bounded by two factors: the occasional inaccuracy of the VLM verifier, and more fundamentally, the quality of the base model's proposals. Our steering method can only select the best option from the provided candidates; if the underlying \textbf{\reasonvla} policy fails to generate any viable action sequences within the batch, verification cannot succeed.

\section{Limitations}
While our method substantially improves robustness, execution consistency, and generalization by verifying reasoning–action alignment at runtime, it is not without limitations.

\textbf{Dependence on Base VLA Quality.} Our runtime steering method selects among candidate actions sampled from the reasoning VLA model. As a result, overall performance is ultimately bounded by the quality and diversity of the base model's proposals.

\textbf{Verifier Reliability.} Our method utilizes the GPT-4o as a VLM verifier for measuring reasoning–action alignment. While we find it as effective or more than specialized Q-value based verifiers, VLMs can produce incorrect judgments, particularly for fine-grained gripper-object interactions in wrist-view images or scenes with severe occlusion.

\textbf{Latency and Compute Overhead.} Although we employ asynchronous verification to mitigate the time of rollout  and obtain lower latency than prior work~\cite{kwok25robomonkey}, our method still introduces additional inference-time latency and cost due to sampling and evaluating multiple candidate sequences. Future work should investigate optimizing the inference infrastructure or use quantization to push the real-time nature of this approach.

\section{Conclusion}

This work addresses what we term the embodied CoT faithfulness gap, where the outcomes described by a VLA’s textual reasoning mismatch its subsequent low-level actions. We introduce a training-free, runtime steering method that enforces this faithfulness by using a VLM to verify multiple candidate action sequences. By selecting the best execution of a plan out of many sampled action sequences, our method converts the policy's natural action diversity from a source of error into a strength to achieve the reasoning-action alignment. Our approach yields performance gains of up to 15\% over strong baselines on challenging OOD and behavior composition tasks, demonstrating that 
verifying what a robot does against what it says is a critical step towards building more robust and trustworthy agents.

\bibliographystyle{bib/IEEEtran}
\bibliography{bib/bibliography}

@inproceedings{mu2025robotwin,
  title={Robotwin: Dual-arm robot benchmark with generative digital twins},
  author={Mu, Yao and Chen, Tianxing and Chen, Zanxin and Peng, Shijia and Lan, Zhiqian and Gao, Zeyu and Liang, Zhixuan and Yu, Qiaojun and Zou, Yude and Xu, Mingkun and others},
  booktitle={Proceedings of the Computer Vision and Pattern Recognition Conference},
  year={2025}
}

@article{alhaija2025cosmos,
  title={Cosmos-transfer1: Conditional world generation with adaptive multimodal control},
  author={Alhaija, Hassan Abu and Alvarez, Jose and Bala, Maciej and Cai, Tiffany and Cao, Tianshi and Cha, Liz and Chen, Joshua and Chen, Mike and Ferroni, Francesco and Fidler, Sanja and others},
  journal={arXiv preprint},
  year={2025}
}

@inproceedings{kostrikovoffline,
  title={Offline Reinforcement Learning with Implicit Q-Learning},
  author={Kostrikov, Ilya and Nair, Ashvin and Levine, Sergey},
  booktitle={International Conference on Learning Representations},
 year = {2021}
}

@inproceedings{open_x_embodiment_rt_x_2023,
  title = {Open {X-E}mbodiment: Robotic Learning Datasets and {RT-X} Models},
  author = {Team, Open X-Embodiment},
  booktitle = {IEEE International Conference on Robotics and Automation (ICRA)},
  year = {2024},
}

@article{zhou2025chatvla,
  title={Chatvla: Unified multimodal understanding and robot control with vision-language-action model},
  author={Zhou, Zhongyi and Zhu, Yichen and Zhu, Minjie and Wen, Junjie and Liu, Ning and Xu, Zhiyuan and Meng, Weibin and Cheng, Ran and Peng, Yaxin and Shen, Chaomin and others},
  journal={arXiv preprint},
  year={2025}
}

@article{wei2022chain,
  title={Chain-of-thought prompting elicits reasoning in large language models},
  author={Wei, Jason and Wang, Xuezhi and Schuurmans, Dale and Bosma, Maarten and Xia, Fei and Chi, Ed and Le, Quoc V and Zhou, Denny and others},
  journal={Advances in neural information processing systems},
  year={2022}
}

@article{comanici2025gemini,
  title={Gemini 2.5: Pushing the frontier with advanced reasoning, multimodality, long context, and next generation agentic capabilities},
  author={Comanici, Gheorghe and Bieber, Eric and Schaekermann, Mike and Pasupat, Ice and Sachdeva, Noveen and Dhillon, Inderjit and Blistein, Marcel and Ram, Ori and Zhang, Dan and Rosen, Evan and others},
  journal={arXiv preprint},
  year={2025}
}

@inproceedings{yang2020multilingual,
  title={Multilingual Universal Sentence Encoder for Semantic Retrieval},
  author={Yang, Yinfei and Cer, Daniel and Ahmad, Amin and Guo, Mandy and Law, Jax and Constant, Noah and Abrego, Gustavo Hernandez and Yuan, Steve and Tar, Chris and Sung, Yun-hsuan and others},
  booktitle={Proceedings of the 58th Annual Meeting of the Association for Computational Linguistics: System Demonstrations},
  year={2020},
}

@article{turpin2023language,
  title={Language models don't always say what they think: Unfaithful explanations in chain-of-thought prompting},
  author={Turpin, Miles and Michael, Julian and Perez, Ethan and Bowman, Samuel},
  journal={Advances in Neural Information Processing Systems},
  year={2023}
}

@article{black2024pi0,
  title={pi 0 : A Vision-Language-Action Flow Model for General Robot Control},
  author={Black, Kevin and Brown, Noah and Driess, Danny and Esmail, Adnan and Equi, Michael and Finn, Chelsea and Fusai, Niccolo and Groom, Lachy and Hausman, Karol and Ichter, Brian and others},
  journal={arXiv preprint},
  year={2024}
}

@inproceedings{nakamoto2024steering,
  author    = {Mitsuhiko Nakamoto and Oier Mees and Aviral Kumar and Sergey Levine},
  title     = {Steering Your Generalists: Improving Robotic Foundation Models via Value Guidance},
  booktitle   = {Conference on Robot Learning (CoRL)},
  year      = {2024},
}

@inproceedings{kwok25robomonkey,
    title={RoboMonkey: Scaling Test-Time Sampling and Verification for Vision-Language-Action Models},
    author={Jacky Kwok and Christopher Agia and Rohan Sinha and Matt Foutter and Shulu Li and Ion Stoica and Azalia Mirhoseini and Marco Pavone},
    booktitle = {Conference on Robot Learning (CoRL)},
    year={2025},
}

@article{lin2025onetwovlaunifiedvisionlanguageactionmodel,
  title={OneTwoVLA: A Unified Vision-Language-Action Model with Adaptive Reasoning},
  author={Fanqi Lin and Ruiqian Nai and Yingdong Hu and Jiacheng You and Junming Zhao and Yang Gao},
  year={2025},
  journal={arXiv preprint}
  
}

@inproceedings{wu2025forewarn,
		title={From Foresight to Forethought: VLM-In-the-Loop Policy Steering via Latent Alignment}, 
		author={Yilin Wu and Ran Tian and Gokul Swamy and Andrea Bajcsy},
		year={2025},
		booktitle={Robotics: Science and System}
  
}

@inproceedings{agiaunpacking,
  title={Unpacking Failure Modes of Generative Policies: Runtime Monitoring of Consistency and Progress},
  author={Agia, Christopher and Sinha, Rohan and Yang, Jingyun and Cao, Ziang and Antonova, Rika and Pavone, Marco and Bohg, Jeannette},
  booktitle={Conference on Robot Learning (CoRL)},
year = {2024}
}

@article{wang2024inference,
  title={Inference-Time Policy Steering through Human Interactions},
  author={Wang, Yanwei and Wang, Lirui and Du, Yilun and Sundaralingam, Balakumar and Yang, Xuning and Chao, Yu-Wei and Perez-D'Arpino, Claudia and Fox, Dieter and Shah, Julie},
  journal={IEEE International Conference on Robotics and Automation (ICRA)},
  year={2025}
}

@article{gao2025taxonomy,
  title={A taxonomy for evaluating generalist robot policies},
  author={Gao, Jensen and Belkhale, Suneel and Dasari, Sudeep and Balakrishna, Ashwin and Shah, Dhruv and Sadigh, Dorsa},
  journal={arXiv preprint},
  year={2025}
}

@article{bjorck2025gr00t,
  title={Gr00t n1: An open foundation model for generalist humanoid robots},
  author={Bjorck, Johan and Casta{\~n}eda, Fernando and Cherniadev, Nikita and Da, Xingye and Ding, Runyu and Fan, Linxi and Fang, Yu and Fox, Dieter and Hu, Fengyuan and Huang, Spencer and others},
  journal={arXiv preprint},
  year={2025}
}

@article{Zawalski24-ecot,
    title={Robotic Control via Embodied Chain-of-Thought Reasoning},
    author={Michał Zawalski and William Chen and Karl Pertsch and Oier Mees and Chelsea Finn and Sergey Levine},
    journal={Conference on Robot Learning (CoRL)},
    year={2024}
}

@article{huang2025thinkact,
  title={ThinkAct: Vision-Language-Action Reasoning via Reinforced Visual Latent Planning},
  author={Huang, Chi-Pin and Wu, Yueh-Hua and Chen, Min-Hung and Wang, Yu-Chiang Frank and Yang, Fu-En},
  journal={arXiv preprint},
  year={2025}
}

@article{snell2024scaling,
  title={Scaling LLM Test-Time Compute Optimally can be More Effective than Scaling Model Parameters},
  author={Snell, Charlie and Lee, Jaehoon and Xu, Kelvin and Kumar, Aviral},
  journal={arXiv e-prints},
  year={2024}
}

@article{intelligence2025pi,
  title={pi 0.5: a Vision-Language-Action Model with Open-World Generalization},
  author={Intelligence, Physical and Black, Kevin and Brown, Noah and Darpinian, James and Dhabalia, Karan and Driess, Danny and Esmail, Adnan and Equi, Michael and Finn, Chelsea and Fusai, Niccolo and others},
  journal={arXiv preprint},
  year={2025}
}

@article{liu2023libero,
  title={Libero: Benchmarking knowledge transfer for lifelong robot learning},
  author={Liu, Bo and Zhu, Yifeng and Gao, Chongkai and Feng, Yihao and Liu, Qiang and Zhu, Yuke and Stone, Peter},
  journal={Advances in Neural Information Processing Systems},
  year={2023}
}

@article{chen25training,
  title={Training Strategies for Efficient Embodied Reasoning},
  author={William Chen and Suneel Belkhale and Suvir Mirchandani and Oier Mees  and Danny Driess and Karl Pertsch and Sergey Levine},
  journal = {Conference on Robot Learning (CoRL)},
  year={2025},
}

@inproceedings{gao2024physically,
  title={Physically grounded vision-language models for robotic manipulation},
  author={Gao, Jensen and Sarkar, Bidipta and Xia, Fei and Xiao, Ted and Wu, Jiajun and Ichter, Brian and Majumdar, Anirudha and Sadigh, Dorsa},
  booktitle={2024 IEEE International Conference on Robotics and Automation (ICRA)},
  year={2024},

}

@article{gupta2025adapting,
  title={Adapting by Analogy: OOD Generalization of Visuomotor Policies via Functional Correspondence},
  author={Gupta, Pranay and Admoni, Henny and Bajcsy, Andrea},
  journal={Conference on Robot Learning (CoRL)},
  year={2025}
}
\clearpage
\appendix
\subsection{Method Details}
\subsubsection{Annotation}
\label{appendix:annotation}
To generate structured reasoning annotations, we employ the video understanding capabilities of Gemini-2.5-Pro~\cite{comanici2025gemini}. Our process begins with the LeRobot-formatted LIBERO-10\footnote{\url{https://huggingface.co/datasets/physical-intelligence/libero}} and LIBERO-90\footnote{\url{https://huggingface.co/datasets/jesbu1/libero_90_lerobot/tree/main}} demonstration datasets. These have been pre-filtered to exclude unsuccessful trajectories and contain agentview image observations corrected via the script provided by the OpenVLA authors.

\begin{figure}[!ht]
    \centering
    \includegraphics[width=\linewidth]{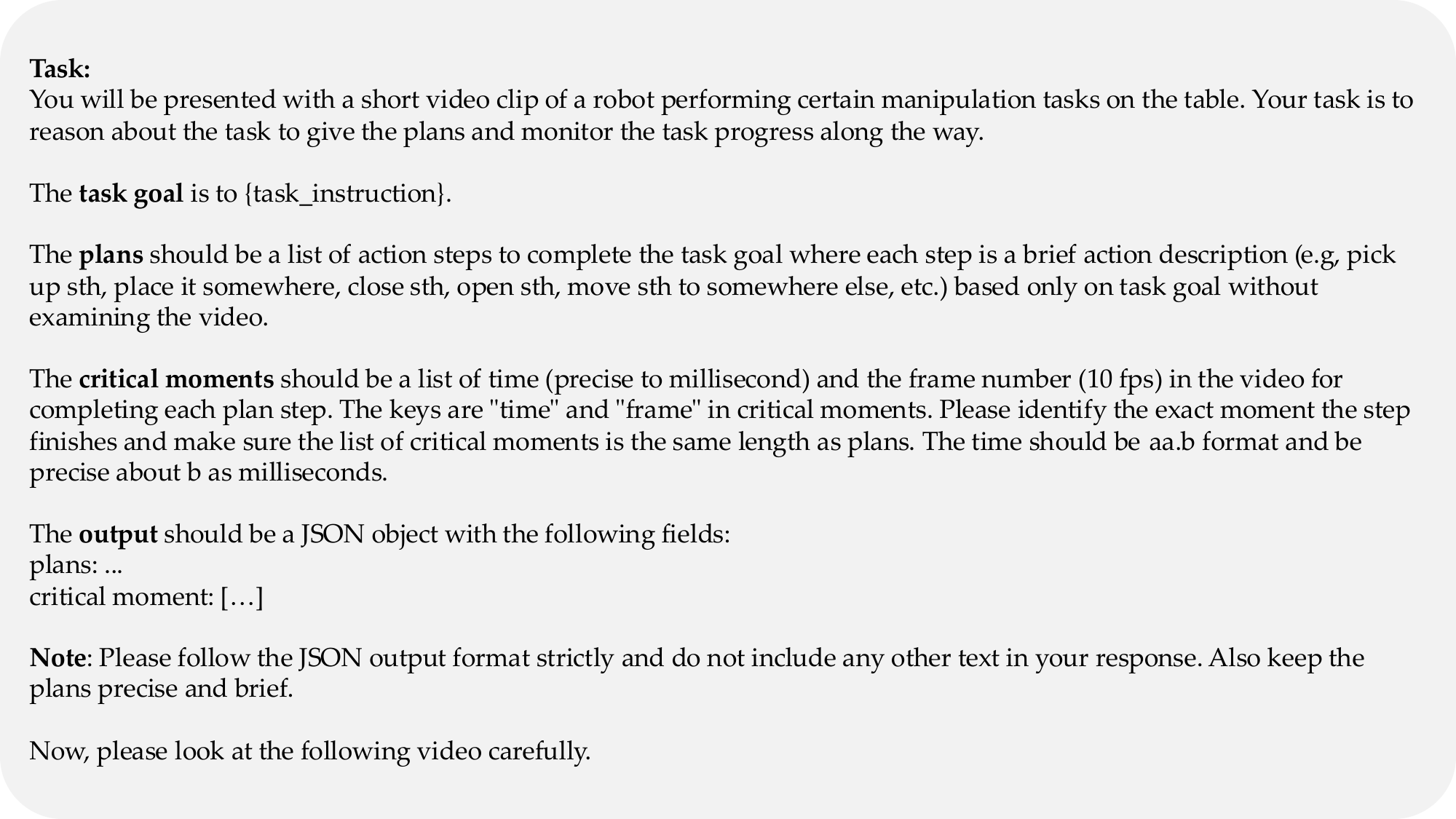}
    \caption{\textbf{Prompt Template for Reasoning Annotations}}
    \label{fig:prompt_annotation}
\end{figure}

For each demonstration episode, we compile the agentview images into a 10 fps video. We then prompt Gemini-2.5-Pro with the task instruction, the corresponding video, and the prompt template shown in Figure~\ref{fig:prompt_annotation}. The model's task is to generate a sequence of textual sub-plans and identify the concluding timestep for each one.

This output is then structured into the following format to serve as the intermediate reasoning signal for our model:

\begin{itemize}
    \item \textit{Plans}: \textless all text plans in the task\textgreater
    \item \textit{What has been done}: \textless completed plans\textgreater
    \item \textit{Now I need to do}: \textless the next text plan to execute\textgreater
\end{itemize}

Since the end of one sub-plan marks the beginning of the next, we use the generated timesteps to align and supervise the model's reasoning generation at the start of each new action sequence.

\subsubsection{Training}
\label{appendix:training}
We follow the recipe in~\cite{lin2025onetwovlaunifiedvisionlanguageactionmodel} to train our base reasoning VLA model \reasonvla. The hyperparameters used for training are listed in Table~\ref{tab:hyperparameter}. The training of \reasonvla takes around 20 hours on 8 A100 GPU or 10 hours on 8 H100 GPU. For our baseline \vla, we use the same hyperparameters used for LIBERO finetuning from openpi repository\footnote{\url{https://github.com/Physical-Intelligence/openpi}}. For learning the Q function in \vgps, we use the original code from the paper~\cite{nakamoto2024steering} and select Implicit Q Learning (IQL)~\cite{kostrikovoffline} algorithm to learn the Q function from offline demonstration dataset LIBERO-100.
\begin{table}[!ht]
    \centering
    \caption{\textbf{Hyperparameters}. The Hyperparameters we use to train and evaluate \reasonvla.}
    \label{tab:hyperparameter}
    \renewcommand{\arraystretch}{1.2}
    \adjustbox{max width=\columnwidth}{
    \begin{tabular}{cl}
        \toprule
        Hyperparameter & Value\\
        \midrule
         Training Batch Size & 128\\
         Validation Batch Size & 64\\
         Training Steps & 30000\\
         Max Token Length & 415 \\
         Text Loss Coefficient & 0.5 \\
         Action Loss Coefficient & 0.5 \\
         Image Size & $(224, 224, 3)$ \\ 
         Prediction Horizon & 16 \\
         Action Horizon & 5 \\
         Image View & Wrist-view and Agent-view \\
        \bottomrule
    \end{tabular}
    }
\end{table}
\subsubsection{Policy Steering}
\label{appendix:steering}
\par\noindent

\para{Hypothesize}
In this stage, \ours generates multiple action sequences in parallel conditioned on the current text plan. The base reasoning model, \reasonvla, uses special tokens (\textless\texttt{act}\textgreater~and~\textless\texttt{think}\textgreater) to switch between planning and acting. We leverage this by using the \textless\texttt{think}\textgreater~token to signal the completion of an action sequence. Specifically, we generate a batch of $K=10$ action sequences concurrently, using KV cache to accelerate inference. As soon as any sequence in the batch generates a \textless\texttt{think}\textgreater~token, it is asynchronously passed to the VLM verifier. While verification runs as a background process, \reasonvla~continues to generate subsequent action tokens for the remaining sequences in the batch. This parallel process terminates as soon as the VLM confirms any one of the proposed sequences as successful.

\para{Predict}
We maintain $K=10$ vectorized environments to simulate the physical outcomes of the hypothesized actions. Each of the $K$ action sequences generated during the \textit{Hypothesize} stage is executed in its own dedicated simulation environment. Following the selection of a successful sequence by the \textit{Verify} stage, we synchronize all parallel environments by setting their states to match the final state of the successful simulation. This ensures that the system proceeds from a single, validated world state before planning the next sub-task.
\begin{table*}
    \centering
    \caption{\textbf{In-Distribution Performance on LIBERO-10}. Success rates (\%) are averaged over 50 trials for each method, conditioned on the training dataset. Our method, \ours, generally outperforms baselines across all training data configurations. For each training dataset and each task, the highest success rate is marked as bold.}
    \label{tab:libero-10-results}
    \renewcommand{\arraystretch}{1.2}
    \begin{tabular}{c cccc cccc cccc}
        \toprule
        \multirow{3.5}{*}{Task ID} & \multicolumn{12}{c}{Success Rate (\%)} \\
        \cmidrule(lr){2-13}
        & \multicolumn{4}{c}{Training Dataset (LIBERO-10)} & \multicolumn{4}{c}{Training Dataset (LIBERO-100-Basket)} & \multicolumn{4}{c}{Training Dataset (LIBERO-100)} \\
        \cmidrule(lr){2-5} \cmidrule(lr){6-9} \cmidrule(lr){10-13}
        & \vla & \vgps & \reasonvla & \ours & \vla & \vgps & \reasonvla & \ours & \vla & \vgps & \reasonvla & \ours \\
        \midrule
        0 & \textbf{96} & 92 & 90 & \textbf{96} & 88 & \textbf{96} & 90 & \textbf{96} & 92 & 88 & 90 & \textbf{100} \\
        1 & \textbf{100} & 98 & 98 & \textbf{100} & \textbf{100} & \textbf{100} & \textbf{100} & \textbf{100} & 94 & 92 & 96 & \textbf{100} \\
        2 & 88 & 94 & 98 & \textbf{100} & 94 & 92 & 80 & \textbf{98} & \textbf{90} & 88 & 62 & 84 \\
        3 & \textbf{100} & \textbf{100} & \textbf{100} & \textbf{100} & 96 & \textbf{98} & 96 & \textbf{98} & 86 & 96 & 98 & \textbf{100} \\
        4 & 76 & 74 & 96 & \textbf{98} & 72 & 78 & 84 & \textbf{88} & 80 & \textbf{98} & \textbf{98} & \textbf{98} \\
        5 & 92 & \textbf{100} & 98 & \textbf{100} & 96 & 98 & 98 & \textbf{100} & 92 & \textbf{98} & \textbf{98} & \textbf{98} \\
        6 & 74 & 80 & 84 & \textbf{98} & 78 & 80 & 80 & \textbf{96} & 82 & 94 & 96 & \textbf{98} \\
        7 & 96 & \textbf{100} & 90 & 98 & 88 & \textbf{92} & \textbf{92} & \textbf{92} & 90 & 90 & 98 & \textbf{100} \\
        8 & 66 & 64 & 66 & \textbf{88} & 44 & 60 & 50 & \textbf{70} & 54 & 56 & 80 & \textbf{90} \\
        9 & 96 & 94 & 98 & \textbf{100} & 90 & 94 & 92 & \textbf{100} & 86 & 88 & 96 & \textbf{100} \\
        \midrule
        Average & 85 & 90 & 92 & \textbf{96} & 85 & 89 & 86 & \textbf{94} & 85 & 87 & 89 & \textbf{97} \\
        \bottomrule
    \end{tabular}
\end{table*}

\begin{figure}
    \centering
    \includegraphics[width=\linewidth]{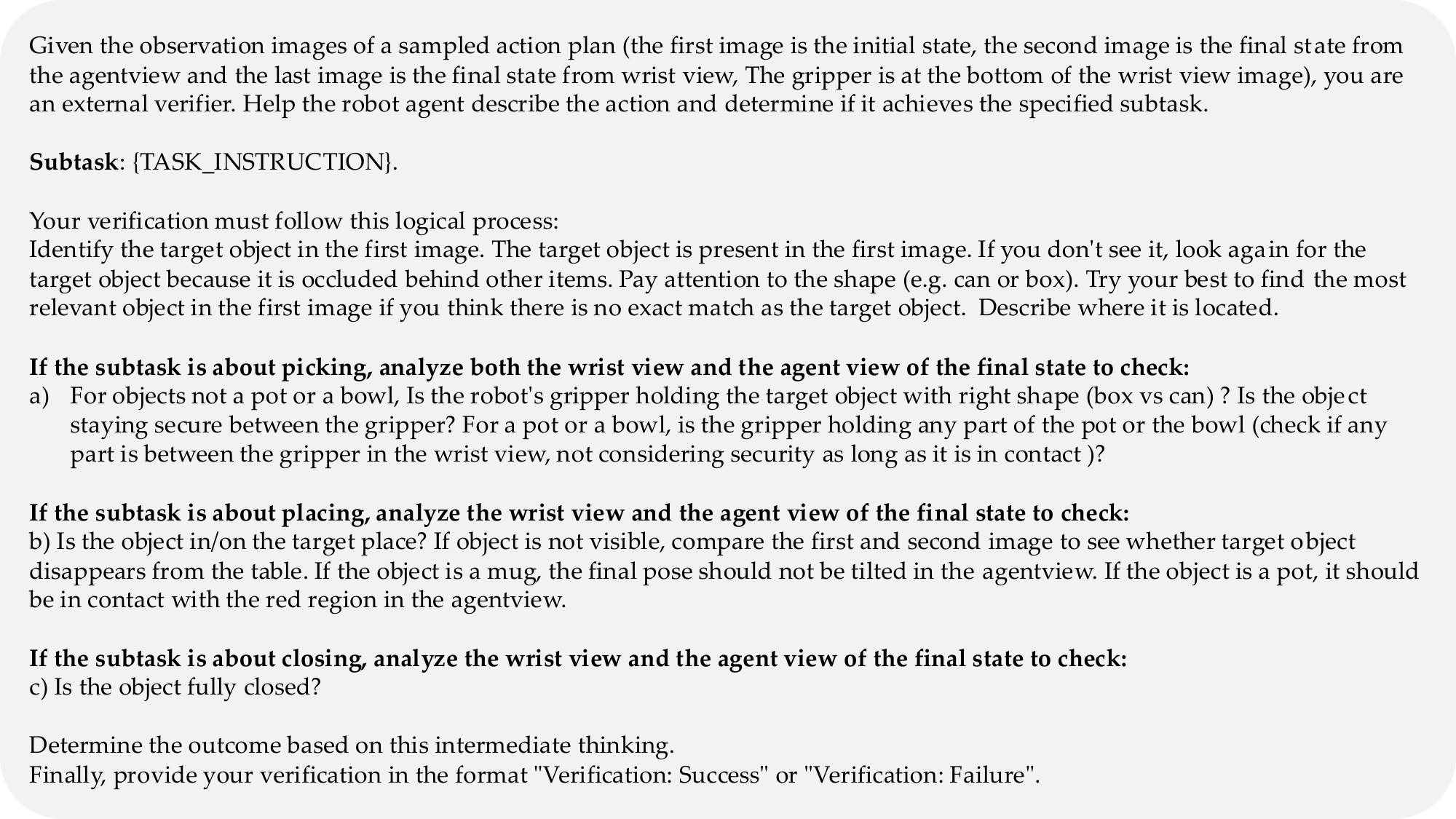}
    \caption{\textbf{Prompt Template for VLM Verification}}
    \label{fig:prompt_verification}
    \vspace{-0.5cm}
\end{figure}

\para{Verify}
We use \textbf{GPT-4o} as a Vision-Language Model (VLM) verifier to approximate a ground-truth reward function. To provide the necessary context, we supply the verifier with the initial agentview image of the task (to establish initial object locations) along with the final agentview and wrist-view images from the completed action sequence. These images are combined with the corresponding textual sub-plan using the prompt template shown in Figure~\ref{fig:prompt_verification}.

We structure the prompt to elicit step-by-step reasoning, guiding the VLM to analyze whether the final state shown in the images matches the objective of the sub-plan. Empirically, we find this structured prompting significantly improves verification accuracy compared to direct queries. However, we note two primary failure modes: the VLM sometimes struggles to correctly assess gripper-object contact, likely because the robot's \texttt{wrist-view} images are out-of-distribution for GPT-4o's training data, and its accuracy decreases in scenes with significant object occlusion.

\subsubsection{Runtime Analysis}
\label{appendix:runtime}
To analyze the computational cost and scalability of our method, we measure the average inference time per step while varying the number of parallel samples ($K$). Fig.~\ref{fig:test_time_scaling} presents this runtime scaling, breaking down the total time into its two primary components: Batch Action Sampling and VLM Verification

As shown in the figure, the total inference time increases with the number of samples. The runtime grows from 147 ms for a single sample to 347 ms for $K=10$ samples.

\begin{itemize}
   \item \textit{Batch Action Sampling}: The cost of Batch Action Sampling scales efficiently and sub-linearly. Its duration increases from 86 ms at $K=1$ to 184 ms at $K=10$. This demonstrates the effectiveness of our batching implementation, where a tenfold increase in samples results in only a \textasciitilde2.1x increase in sampling time.

    \item \textit{VLM Verification}: In contrast, the time spent on VLM Verification scales more sharply, especially at higher sample counts. This component's average cost per step grows from 61 ms for one sample to 163 ms for ten samples. Although the VLM verification runs in the background, each query still takes 7-10 seconds for GPT-4o to generate the answer because of the thinking mode.
\end{itemize}

The results indicate a trade-off between potential performance gains from more samples and the resulting inference latency. While our system supports parallel sampling efficiently, the VLM verification step becomes a more significant bottleneck as the number of samples increases.

\subsection{Benchmark Details}
\label{appendix:benchmark}

\subsubsection{In-distribution Tasks}
\label{appendix:id_task}
We evaluate on ten long-horizon manipulation tasks from the LIBERO-10 benchmark~\cite{liu2023libero} as our in-distribution test set. These tasks are well-suited for Vision Language Action Models (VLAs) with intermediate reasoning capabilities due to their long-horizon nature. The instructions for each task are listed in Table~\ref{tab:libero-10-instruction}.

\begin{table}
    \centering
    \caption{\textbf{Task Instructions for LIBERO-10}. The ten long-horizon tasks and their corresponding language instructions.}
    \label{tab:libero-10-instruction}
    \renewcommand{\arraystretch}{1.2}
    \adjustbox{max width=\columnwidth}{
    \begin{tabular}{cl}
        \toprule
        Task ID & Task Instruction \\
        \midrule
        0 & put both the alphabet soup and the tomato sauce in the basket. \\
        1 & put both the cream cheese box and the butter in the basket. \\
        2 & turn on the stove and put the moka pot on it. \\
        3 & put the black bowl in the bottom drawer of the cabinet and close it. \\
        4 & put the white mug on the left plate and put the yellow and white mug on the right plate. \\
        5 & pick up the book and place it in the back compartment of the caddy. \\
        6 & put the white mug on the plate and put the chocolate pudding to the right of the plate. \\
        7 & put both the alphabet soup and the cream cheese box in the basket. \\
        8 & put both moka pots on the stove. \\
        9 & put the yellow and white mug in the microwave and close it. \\
        \bottomrule
    \end{tabular}
    }
\end{table}

We evaluate all methods on these tasks after training them on three different datasets. The per-task success rates are detailed in Table~\ref{tab:libero-10-results}, and the average performance is reported in Figure~\ref{fig:exp_compose} of the main paper.

\begin{figure*}
    \centering
    \includegraphics[width=0.9\linewidth]{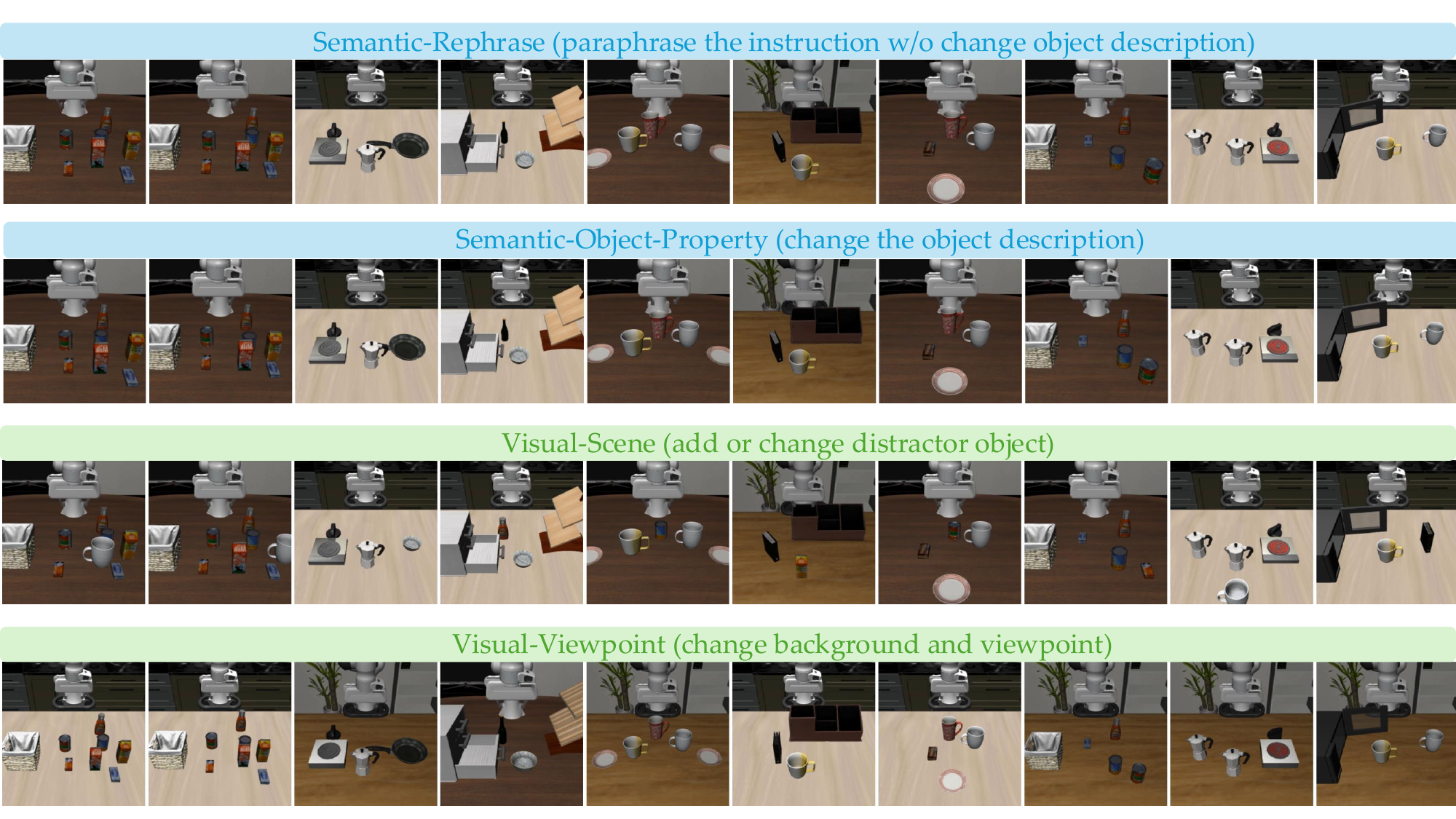}
    \caption{\textbf{Visualizations of Initial States for OOD Tasks}. The top row shows the semantic OOD shift of Lang-Rephrase from LIBERO-10. The second row corresponds to semantic shifts with Lang-Object-Property. These two rows are the same as LIBERO-10 since they have no visual changes. The third row shows the \emph{Visual-Scene} shift with object substitutions or additions. The bottom row shows the \emph{Visual-Viewpoint} shift with a new background and camera angle.}
    \vspace{-0.5cm}
    \label{fig:ood_initial_state}
\end{figure*}
\subsubsection{Out-of-Distribution (OOD) Tasks}
\label{appendix:ood_task}
Following~\cite{gao2024physically}, we extend the LIBERO-10 benchmark with four out-of-distribution (OOD) variations, categorized as semantic or visual shifts.

\begin{itemize}
    \item \textbf{Semantic OOD}: The task instruction is altered while the scene remains the same.
    \begin{itemize}
        \item \emph{Lang-Rephrase}: The instruction is paraphrased using different verbs and sentence structures, but the object descriptions remain the same. The new instructions are in Table~\ref{tab:libero-10-semantic-1-instruction}.
        \item \emph{Lang-Object-Property}: The object descriptions are replaced with different but equivalent references (e.g., "yellow and white mug" becomes "middle mug"). The new instructions are in Table~\ref{tab:libero-10-semantic-2-instruction}.
    \end{itemize}
    \item \textbf{Visual OOD}: The visual properties of the scene are changed while the instruction remains the same.
    \begin{itemize}
        \item \emph{Visual-Scene}: A non-target object is replaced, or a distractor object is added to the scene (e.g., a wine bottle is replaced with a ketchup bottle). See the third row of Figure~\ref{fig:ood_initial_state} for visualizations.
        \item \emph{Visual-Viewpoint}: The background and camera pose are changed. We randomly select one of two alternate backgrounds available in LIBERO (Kitchen, Living Room, Study Room) that was not used in the original task. The camera pose is adjusted to keep all objects in view. See the fourth row of Figure~\ref{fig:ood_initial_state} for visualizations.
    \end{itemize}
\end{itemize}

We evaluate all methods trained on the LIBERO-100 dataset against these four OOD settings, as models trained on larger, more diverse data are expected to be more robust and LIBERO-100 is our largest dataset. Per-task results are presented in Table~\ref{tab:libero-10-ood-results}, with average performance summarized in Figure~\ref{fig:ood-generalization}.
\begin{table}[!ht]
    \centering
    \caption{\textbf{Task Instructions for Semantic OOD (Lang-Rephrase)}. The original instructions are paraphrased, altering sentence structure and verbs while keeping object descriptions intact.}
    \label{tab:libero-10-semantic-1-instruction}
    \renewcommand{\arraystretch}{1.2}
    \adjustbox{max width=\columnwidth}{
    \begin{tabular}{cl}
        \toprule
        Task ID & Task Instruction \\
        \midrule
        0 & place the alphabet soup in the basket and the tomato sauce as well. \\
        1 & pick up both the cream cheese box and the butter and place them in the basket. \\
        2 & switch the stove on and set the moka pot on the stove. \\
        3 & move the black bowl to the bottom drawer and close the drawer of the cabinet. \\
        4 & place two mugs on the plates, left one is the white mug and the right one is the yellow and white mug. \\
        5 & grab the standing book and transfer it to the back compartment of the caddy. \\
        6 & set the white mug on the plate with chocolate pudding to be placed to the right. \\
        7 & put both the can of soup and the box of cheese in the basket. \\
        8 & transfer both moka pots from the table to the stove. \\
        9 & place the yellow and white mug inside the microwave, then shut the door. \\
        \bottomrule
    \end{tabular}
    }
\end{table}

\begin{table}[!ht]
    \centering
    \caption{\textbf{Task Instructions for Semantic OOD (Lang-Object-Property)}. The way objects are described is changed, while the core task and sentence structure remain similar.}
    \label{tab:libero-10-semantic-2-instruction}
    \renewcommand{\arraystretch}{1.2}
    \adjustbox{max width=\columnwidth}{
    \begin{tabular}{cl}
        \toprule
        Task ID & Task Instruction \\
        \midrule
        0 & put both the can of soup and can of sauce in the basket. \\
        1 & put both the box of cheese and the box of butter in the basket. \\
        2 & turn on the cooktop and place the moka machine on it. \\
        3 & put the middle bowl to the lowest drawer and close it. \\
        4 & put the pure white cup on the left plate and put the other one with the yellow handle on the right plate. \\
        5 & pick up the right book and put it in the rear part of the caddy. \\
        6 & put the pure white cup on the middle plate and put the brown chocolate to the right of the plate. \\
        7 & move the two objects, alphabet soup and cream cheese box, to the basket. \\
        8 & put both the moka coffee makers on the cooktop. \\
        9 & put the middle mug inside the microwave and close the door. \\
        \bottomrule
    \end{tabular}
    }
\end{table}
\begin{table*}[!ht]
    \centering
    \caption{\textbf{OOD Performance on LIBERO-10 Variants}. Success rates (\%) are averaged over 50 trials for each method trained on the LIBERO-100 dataset. Our method, \ours, demonstrates stronger robustness to both semantic and visual OOD shifts. For each OOD shift and each task, the highest success rate is marked as bold.}
    \label{tab:libero-10-ood-results}
    \renewcommand{\arraystretch}{1.2}
    \adjustbox{max width=\linewidth}{
    \begin{tabular}{c cccc cccc cccc cccc}
        \toprule
        \multirow{3.5}{*}{Task ID} & \multicolumn{16}{c}{Success Rate (\%)} \\
        \cmidrule(lr){2-17}
        & \multicolumn{4}{c}{Semantic OOD (Lang-Rephrase)} & \multicolumn{4}{c}{Semantic OOD (Lang-Object-Property)} & \multicolumn{4}{c}{Visual OOD (Visual-Scene)} & \multicolumn{4}{c}{Visual OOD (Visual-Viewpoint)} \\
        \cmidrule(lr){2-5} \cmidrule(lr){6-9} \cmidrule(lr){10-13} \cmidrule(lr){14-17}
        & \vla & \vgps & \reasonvla & \ours & \vla & \vgps & \reasonvla & \ours & \vla & \vgps & \reasonvla & \ours & \vla & \vgps & \reasonvla & \ours \\
        \midrule
        0 & 86 & 92 & 92 & \textbf{98} & 72 & 80 & 90 & \textbf{98} & \textbf{96} & 88 & \textbf{96} & \textbf{96} & 52 & 30 & 12 & \textbf{68} \\
        1 & 92 & 96 & 98 & \textbf{100} & 96 & 96 & 98 & \textbf{100} & \textbf{100} & 98 & \textbf{100} & \textbf{100} & 34 & 36 & 72 & \textbf{84} \\
        2 & 32 & 32 & 82 & \textbf{96} & 66 & 70 & 54 & \textbf{72} & \textbf{94} & 92 & 64 & 90 & 4 & 0 & 8 & \textbf{10} \\
        3 & 86 & 88 & 96 & \textbf{98} & 90 & 88 & 94 & \textbf{98} & 94 & 96 & \textbf{100} & \textbf{100} & \textbf{72} & 66 & 2 & 32 \\
        4 & 32 & 14 & 78 & \textbf{84} & 70 & 74 & 84 & \textbf{90} & 90 & 70 & 96 & \textbf{98} & 0 & 8 & 24 & \textbf{40} \\
        5 & 96 & 94 & \textbf{100} & \textbf{100} & 18 & \textbf{100} & 76 & 96 & 92 & 96 & 98 & \textbf{100} & 30 & 26 & 58 & \textbf{94} \\
        6 & 86 & 88 & 76 & \textbf{96} & 86 & \textbf{88} & 76 & 84 & 88 & \textbf{98} & 96 & 96 & 34 & 32 & 30 & \textbf{44} \\
        7 & 86 & \textbf{92} & 74 & 82 & 90 & 90 & 94 & \textbf{98} & 88 & 94 & 94 & \textbf{100} & 20 & 14 & 24 & \textbf{48} \\
        8 & 62 & 44 & 78 & \textbf{98} & 58 & 54 & 68 & \textbf{84} & 4 & 4 & 68 & \textbf{98} & 10 & 6 & 12 & \textbf{20} \\
        9 & 74 & 74 & 88 & \textbf{98} & 78 & 70 & 80 & \textbf{92} & 92 & 92 & 96 & \textbf{100} & \textbf{27} & 12 & 8 & 22 \\
        \midrule
        Average & 73 & 71 & 86 & \textbf{95} & 72 & 81 & 81 & \textbf{91} & 84 & 83 & 91 & \textbf{98} & 28 & 23 & 25 & \textbf{45} \\
        \bottomrule
    \end{tabular}
    }
\end{table*}
\begin{table*}
    \centering
    \caption{\textbf{Compositional Generalization Performance}. Success rates (\%) are averaged over 50 trials. As the training dataset size and diversity increase from LIBERO-10 to LIBERO-100, our method's performance advantage over baselines becomes more pronounced. Entries marked with `---` are not applicable for that training dataset. For each training dataset and each task, the highest success rate is marked as bold.}
    \label{tab:libero-10-compose-result}
    \renewcommand{\arraystretch}{1.2}
    \begin{tabular}{c cccc cccc cccc}
        \toprule
        \multirow{3.5}{*}{Task ID} & \multicolumn{12}{c}{Success Rate (\%)} \\
        \cmidrule(lr){2-13}
        & \multicolumn{4}{c}{Training Dataset (LIBERO-10)} & \multicolumn{4}{c}{Training Dataset (LIBERO-100-Basket)} & \multicolumn{4}{c}{Training Dataset (LIBERO-100)} \\
        \cmidrule(lr){2-5} \cmidrule(lr){6-9} \cmidrule(lr){10-13}
        & \vla & \vgps & \reasonvla & \ours & \vla & \vgps & \reasonvla & \ours & \vla & \vgps & \reasonvla & \ours \\
        \midrule
        0 & 28 & 32 & \textbf{34} & 32 & 4 & 8 & 6 & \textbf{10} & 26 & 46 & 30 & \textbf{82} \\
        1 & 0 & \textbf{2} & \textbf{2} & 0 & 0 & 0 & 0 & \textbf{4} & 20 & 6 & 42 & \textbf{60} \\
        2 & --- & --- & --- & --- & 22 & 28 & \textbf{74} & 70 & 22 & 18 & 76 & \textbf{88} \\
        3 & --- & --- & --- & --- & 2 & 2 & 18 & \textbf{28} & 14 & \textbf{18} & 2 & 4 \\
        4 & --- & --- & --- & --- & 0 & 2 & 6 & \textbf{12} & 0 & 2 & 46 & \textbf{60} \\
        5 & --- & --- & --- & --- & 0 & 0 & \textbf{6} & 4 & 6 & 2 & 48 & \textbf{72} \\
        6 & --- & --- & --- & --- & 2 & 2 & \textbf{14} & 10 & 24 & \textbf{34} & 0 & 4 \\
        7 & --- & --- & --- & --- & 0 & 2 & 4 & \textbf{6} & 30 & 24 & 50 & \textbf{74} \\
        8 & --- & --- & --- & --- & 68 & 70 & 82 & \textbf{90} & 54 & 34 & 94 & \textbf{96} \\
        9 & --- & --- & --- & --- & --- & --- & --- & --- & 10 & 12 & 14 & \textbf{20} \\
        10 & --- & --- & --- & --- & --- & --- & --- & --- & 2 & 2 & 14 & \textbf{38} \\
        11 & --- & --- & --- & --- & --- & --- & --- & --- & 2 & 0 & 4 & \textbf{14} \\
        12 & --- & --- & --- & --- & --- & --- & --- & --- & 0 & 4 & \textbf{80} & 78 \\
        \midrule
        Average & 14 & 17 & \textbf{18} & 16 & 11 & 13 & 23 & \textbf{26} & 16 & 16 & 38 & \textbf{53} \\
        \bottomrule
    \end{tabular}
\end{table*}

\subsubsection{Behavior Composition Tasks}
\label{appendix:generalization_task}
To further test generalization, we introduce a suite of novel \emph{behavior composition} tasks. These tasks require reusing skills learned from training demonstrations in combinations not seen during training. For instance, if a model has seen demonstrations for "put both the alphabet soup and the tomato sauce in the basket" and "put the orange juice in the basket," a compositional task might be "put both the orange juice and the tomato sauce in the basket."

The set of compositional tasks depends on the training dataset. For models trained on \emph{LIBERO-10} dataset, we create two new tasks by recombining objects to put in the basket (Tasks 0-1 in Table~\ref{tab:libero-10-compose-instruction}). For \emph{LIBERO-100-Basket}, which includes more pick-and-place examples for basket , we use a larger set of nine tasks (Tasks 0-8). For the most diverse \emph{LIBERO-100} dataset, we create the full suite of 13 tasks including more objects like mugs and wine bottle. The complete list of tasks is provided in Table~\ref{tab:libero-10-compose-instruction}.

We evaluate all methods on their corresponding compositional task suite. Per-task results are in Table~\ref{tab:libero-10-compose-result}, and average performance is reported in Figure~\ref{fig:exp_compose} of the main paper.
\begin{table}
    \centering
    \caption{\textbf{Task Instructions for Behavior Composition Tasks}. These new tasks require combinations of skills that were not observed during training.}
    \label{tab:libero-10-compose-instruction}
    \renewcommand{\arraystretch}{1.2}
    \adjustbox{max width=\columnwidth}{
    \begin{tabular}{cl}
        \toprule
        Task ID & Task Instruction \\
        \midrule
        0 & put both the cream cheese and the tomato sauce in the basket. \\
        1 & put both the alphabet soup and the butter in the basket. \\
        2 & put both the orange juice and the tomato sauce in the basket. \\
        3 & put both the milk and the tomato sauce in the basket. \\
        4 & put both the orange juice and the butter in the basket. \\
        5 & put both the tomato sauce and the butter in the basket. \\
        6 & put both the milk and the butter in the basket. \\
        7 & put both the ketchup and the cream cheese in the basket. \\
        8 & put both the tomato sauce and the cream cheese box in the basket. \\
        9 & put the wine bottle in the bottom drawer of the cabinet and close it. \\
        10 & put the red mug on the plate and put the chocolate pudding to the right of the plate. \\
        11 & put the red mug on the plate and put the chocolate pudding to the left of the plate. \\
        12 & put the red mug on the left plate and put the yellow and white mug on the right plate. \\
        \bottomrule
    \end{tabular}
    }
\end{table}

\end{document}